\documentclass[sigconf]{acmart}
\usepackage{dsfont}
\usepackage{xspace}
\usepackage{amsthm}
\usepackage{amsmath}
\usepackage{enumitem}
\usepackage{booktabs}
\usepackage{multirow}
\usepackage{graphicx}
\usepackage{bbm}
\usepackage[normalem]{ulem}
\useunder{\uline}{\ul}{}
\usepackage{float}
\usepackage{caption}
\usepackage{subcaption}
\usepackage{balance}
\usepackage{threeparttable}
\usepackage{multicol}
\usepackage{multirow}
\usepackage[export]{adjustbox}
\usepackage{bm}
\usepackage{pifont}
\usepackage{threeparttable}
\usepackage{adjustbox}
\usepackage{array}
\newcommand{\modelname}{\textsf{CDDRec}\xspace}

\begin{document}

\title{Conditional Denoising Diffusion for Sequential Recommendation}

\author{Yu Wang}
\affiliation{%
  \institution{University of Illinois Chicago}
 \city{Chicago}
  \country{United States}
}  
\email{ywang617@uic.edu}

\author{Zhiwei Liu}
\affiliation{%
  \institution{Salesforce}
 \city{San Francisco}
  \country{United States}
}
\email{zhiweiliu@salesforce.com}

\author{Liangwei Yang}
\affiliation{%
  \institution{University of Illinois Chicago}
 \city{Chicago}
  \country{United States}
}
\email{lyang84@uic.edu}
\author{Philip S. Yu}
\affiliation{%
  \institution{University of Illinois Chicago}
 \city{Chicago}
  \country{United States}
}
\email{psyu@uic.edu}

\renewcommand{\shortauthors}{Yu Wang et al.}

\begin{abstract}

Generative models have attracted significant interest due to their ability to handle uncertainty by learning the inherent data distributions. However, two prominent generative models, namely Generative Adversarial Networks (GANs) and Variational AutoEncoders (VAEs), exhibit challenges that impede achieving optimal performance in sequential recommendation tasks. Specifically, GANs suffer from unstable optimization, while VAEs are prone to posterior collapse and over-smoothed generations. The sparse and noisy nature of sequential recommendation further exacerbates these issues.

In response to these limitations, we present a conditional denoising diffusion model, which includes a sequence encoder, a cross-attentive denoising decoder, and a step-wise diffuser. This approach streamlines the optimization and generation process by dividing it into easier and tractable steps in a conditional autoregressive manner. Furthermore, we introduce a novel optimization schema that incorporates both cross-divergence loss and contrastive loss. This novel training schema enables the model to generate high-quality sequence/item representations and meanwhile precluding collapse. We conducted comprehensive experiments on four benchmark datasets, and the superior performance achieved by our model attests to its efficacy. 

\end{abstract}


\maketitle

\section{Introduction}

Sequential Recommendation~\cite{SASRec, TiSASRec, ACVAE, wang2022contrastvae, liu2021augmenting, seq2bubble} suggests items to users based on their previous interactions such as purchases, clicks, or ratings.
It has been intensively investigated because of its scalability and efficacy in capturing user temporal trends. 
Recent research in sequential recommendation focuses on attention-based methods for their promising results. SASRec~\cite{SASRec} and Bert4Rec~\cite{Bert4Rec} are early attempts that utilize the attention-based transformer structure in the sequential recommendation. Further improvements in item representation quality are made by DuoRec~\cite{DuoRec} and CBiT~\cite{CBiT}, which incorporated contrastive learning to make items more distinguishable without compromising ranking capabilities. Though attention-based methods demonstrate their effectiveness, they still suffer from performance degradation if there are noisy interactions, like random clicks in sequences.
These noises lead to biased ranking results, especially for short sequences and cold-start items~\cite{wang2022contrastvae}. 

In this sense, generative models~\cite{vae,gan} are running into the spotlight. 
They provide solutions by learning the underlying distribution and estimating the uncertainty within the data, which improves their robustness to noises~\cite{ratzlaff2019hypergan, mohamed2016learning, grover2019uncertainty, wang2022contrastvae}.
Two prevalent categories of generative models are Generative Adversarial Networks (GANs)~\cite{gan} and Variational AutoEncoders (VAEs)~\cite{vae}. 
Compared with GAN-based methods, VAE-based methods have more stable optimization, which thus have garnered more attention in the sequential recommendation.
% VAE-based methods, which offer stable optimization compared to GANs, have gained popularity in the sequential recommendation domain.
ACVAE~\cite{ACVAE} incorporates adversarial training into the VAE framework, while ContrastVAE~\cite{wang2022contrastvae} integrates contrastive learning into the VAE to address posterior collapse issues, showing promising improvements compared to conventional recommender systems.

Despite the success of generative models, there are limitations that impede these models from achieving enhanced performance. 
First, it is hard for generative models to well characterize the distribution of discrete sequences. 
Specifically, the optimization of GAN-based methods is unstable~\cite{miyato2018spectral, cao2019multi, lee2021vitgan, salimans2016improved}, while the VAE-based methods suffer from posterior collapse issues~\cite{InfoVAE, MultVAE, takida2021preventing, tang2021exploring}.
Secondly, the sparsity of sequential recommendation aggravates the difficulty in training generative models. 
For example, in short sequences, the portion of noisy interactions is relatively larger than in long sequences. 
Hence, the generative models are unable to capture the true distribution of the data. 
Moreover, we observe that generative models especially for those VAE-based models tend to yield over-smoothing results~\cite{nowozin2016f, rezende2018taming}.
Specifically, generative models are prone to yield similar ranking scores on candidate items, which results from the incorporation of uncertainty. 
Though obtaining good performance when considering a set of top-ranking items, existing generative models are incapable of ranking those items in accurate positions as their scores are rather close.

To address these issues, we can divide the difficult and unstable learning procedure into multiple simple tasks and split the generation procedure into multi-steps. Intuitively, if the noisy interpolation has relatively more impact on short sequences, we can simulate the smaller level of noise for each step, thus simplifying the learning task.
If the one-step generated ranking prediction is too smooth, we can learn the small transition for each step and gradually generate the high-fidelity ranking predictions. For these desiderata, we resort to denoising diffusion models (DDMs) for solutions. These models~\cite{ddpm, dalle2,gao2022difformer} can be regarded as the extended VAEs that overcome both posterior collapse and over-smoothing issues through the fine-grained multi-step generation and inherit the nature of the stable and efficient optimization of VAEs. Specifically, DDMs consist of two phases: diffusion and denoising. The diffusion step collapses original inputs into Gaussian noise through gradual noise addition, while the denoising step generates original input from Gaussian noise by removing the noise step-by-step. Generally, the diffuser of DDMs maps the input to the Gaussian Noise gradually and the denoiser reconstructs such intermediate states. As the noise added to the input from a diffuser is scheduled and relatively small, the denoiser can be aware of such noise and learn the step-wise transition to reconstruct the original information from Gaussian Noise. Furthermore, due to such multi-step generations, the decoder is able to learn fine-grained intermediate transitions and overcome the over-smoothing issues of VAE from the one-step generation paradigm.  

However, it is rather difficult to incorporate such a learning paradigm into the sequential recommendation. There are only a few successful examples in discrete tasks, such as text generation~\cite{diffusion-lm, gao2022difformer, gong2022diffuseq}.
%original objective collapse
One primary reason is that traditional diffusion models are designed for continuous spaces like image generation, where input features are fixed and contain substantial information. In contrast, sequential recommendation scenarios involve item input information that is randomly initialized based on item IDs. Without a carefully designed approach that takes into account the characteristics of DDMs, performance in discrete spaces could be adversely affected by representation collapse~\cite{gao2022difformer, diffusion-lm, gong2022diffuseq}. 
Another factor that hinders the successful implementation of DDMs in sequential recommendation is its original purpose: reconstructing the corrupted original information. In the context of sequential recommendation, the desired outcome is the generation of high-quality sequence/item representations reflecting user preferences, enabling the prediction of the next possible items based on prior interaction records. Merely reconstructing original item representations within a sequence (Gaussian Noise vector from the beginning) could exacerbate the collapse issues of DDMs for the sequential recommendation. 
Furthermore, the generation process of DDMs is unconditional and non-autoregressive, which is inappropriate for the sequential recommendation. Specifically, the generation of DDM starts from the randomly initialized Gaussian Noise, which is uncontrollable. For sequential recommendation, as the autoregressive generation has been proven to be effective~\cite{SASRec, wang2022contrastvae}, we expect the model could generate the next item engagement representation that is conditioned on previous interactions, i.e., conditionally generate sequence representations autoregressively. 

To tackle these challenges, we propose conditional denoising diffusion models for sequential recommendation (\modelname) that include a sequence encoder, cross-attentive conditional denoising decoder, and step-wise diffuser. 
The sequence encoder encodes sequence representations from interacted item embeddings, which is used as the conditioned information for a step-wise generation. 
The objective of the conditional denoising decoder is to generate high-quality sequence representations by removing the noise of sequences step-by-step. To make the denoising decoder aware of each denoising step, we adopt the cross-attention mechanism with the denoising step as the query input. 
To boost the generation performance, previous methods choose the self-condition strategy, which makes the generation process condition on the output of the previous generation step~\cite{gao2022difformer}. It can be regarded as adding a residual connection to the long dependence chain. In the conditional generation scenario, we make the denoising decoder estimate the noise of every step directly conditional on the output of the sequence encoder. 
The step-wise diffuser introduces noise into target sequence representations to construct corrupted targets and simulate the small step-wise noise in the sequences. 
We also introduce a cross-divergence loss based on DDM's original reconstruction loss, enabling the model to construct high-fidelity sequence/item representations and be aware of user preferences while preventing learning collapse. Furthermore, we introduce In-view and Cross-view contrastive optimization to ensure the model predicts a consistent output given the noise interpolation.

Our contribution can be summarized as follows:
\vspace{-2mm}
\begin{itemize}
    \item To the best of our knowledge, we propose the novel conditional denoise diffusion models for sequential recommendation \modelname in the conditional autoregressive generation paradigm.
    \item We designed the optimization schema to equip the \modelname with the ranking capacity for sequential recommendation and prevent it from representation collapse.
    \item We conduct comprehensive experiments on sequential recommendation dataset, the substantial improvement on all metrics through four datasets indicates the effectiveness of \modelname. We also conduct ablation studies to further examine the effectiveness of each key design. 
\end{itemize}

\section{Related Work}
\subsection{Generative Models for Sequential Recommendation}
Generative models have been extensively studied and applied in the sequential recommendation. Variational AutoEncoder (VAE) is a popular generative model that learns item latent representations by first encoding them into a latent space and then reconstructing the original data sample from this latent space. \citet{MultVAE} applied VAEs to recommendation tasks by assuming that the interacted items of a user follow the multinomial distribution and optimizing the VAE by maximizing the reconstruction likelihood. \citet{SVAE} further improved upon this method by factorizing the joint distribution of a user interaction sequence in an autoregressive manner. \citet{VSAN} used self-attention layers to implement the encoder and decoder of the VAE. \citet{wang2022contrastvae} proposed the ContrastVAE, which incorporates contrastive learning into the VAE framework and utilizes the contrastELBO from a multi-view perspective. Generative Adversarial Network (GAN) is another widely used generative model. GANs optimize a generator and a discriminator simultaneously. The discriminator distinguishes between samples generated by the generator and those from the real dataset, while the generator generates indistinguishable samples that maximize the prediction error of the discriminator. \citet{ren2020sequential} utilized a generator to generate sequence representations that predict the next-item ranking score and introduced a multi-factor specific discriminator to estimate the rationality of the generated embedding with respect to each item factor, guiding the generation process. \citet{ACVAE} introduced adversarial training to VAEs, enabling the model to learn more diverse and informative item representations.

\subsection{Denoise Diffusion models for discrete tasks}
Denoise Diffusion Models (DDMs) have shown great success in continuous spaces, such as image generation~\cite{ddpm, dalle2, nichol2021improved, huang2021variational} and audio generation~\cite{kong2020diffwave, lee2021priorgrad, kingma2021variational, lam2022bddm}. Recently, several attempts have been made to apply DDMs in discrete spaces, such as text generation. SUNDAE~\cite{SUNDAE} is one of the pioneers that uses DDMs for text generation. They introduce a step-unrolled denoising autoencoder that reconstructs corrupted sequences in a non-autoregressive manner. Diffusion-LM~\cite{diffusion-lm} gradually reconstructs word vectors from Gaussian noise guided by attribute classifiers and introduces a rounding process that maps continuous word embeddings to discrete words. DiffSeq~\cite{gong2022diffuseq} introduces a forward process with partial noise that uses the question of a dialog as the uncorrupted part and the answer of the dialog as the corrupted part and adds partial noise to the answer part during the forward pass. The backward pass reconstructs the answer in a non-autoregressive way. While the denoising process in DDMs realizes conditional generation, this methodology is not directly applicable for sequential recommendation tasks, as there is significant overlap when regarding historical interactions as questions and one-step-shifted interactions as answers. Additionally, such models may be unaware of rankings among items and user preferences. TimeGrad~\cite{TimeGrad} is one attempt to incorporate DDMs into the conditional autoregressive generation paradigm, utilizing an RNN to encode previous sequences and applying multi-step diffusion/denoising over the latent states of the RNN.

\section{Preliminary}\label{sec:prelim}

\paragraph{ Denoising Diffusion Models (DDMs)} DDMs have demonstrated considerable effectiveness in generating high-quality data for continuous tasks, encompassing areas such as computer vision~\cite{ddpm, dalle2, nichol2021improved, huang2021variational}  and audio generation~\cite{kong2020diffwave, lee2021priorgrad, kingma2021variational, lam2022bddm}. These models belong to the family of likelihood-based generative models. However, in contrast to the variational autoencoder (a notable likelihood-based generative model) which directly produces data from latent embeddings in a single step, DDMs divide the generation process into multiple stages by restoring corrupted data. It is the parameterized Markov chain that progressively introduces noise until the original data is reduced to Gaussian noise during the diffusion phase. Conversely, the reverse denoising phase is learned to recover the corrupted data at each stage. Upon mastering the reverse denoising phase, the model is adept at denoising data from randomly sampled Gaussian noise in a stepwise manner. This multi-step refinement approach simplifies the learning process and facilitates high-fidelity generation. It is important to note that data corruption occurs by adding a Gaussian noise scale to the original data according to a noise schedule strategy, specifically, incorporating fixed Gaussian noise at each stage. Given the similarities between DDMs and VAEs as likelihood-based generative models, the training objectives for both models begin with optimizing the evidence lower bound (ELBO) for minimizing the negative log-likelihood:
\vspace{-2mm}
\begin{equation}\label{eq:L_DDM}
    \mathcal{L}_{DDM} = -\mathbb{E}_{q(x_0)}\log p_{\theta}(x_0) \leq \mathbb{E}_{q(x_0, x_{1:T})}[\log \frac{q(x_{1:T}|x_0)}{p_{\theta}(x_0,x_{1:T})}],
\end{equation}
\vspace{-2mm}
\begin{equation}
    \mathcal{L}_{VAE} = -\mathbb{E}_{q(x_0)}\log p_{\theta}(x_0) \leq \mathbb{E}_{q(x_0,z)}[\log \frac{q(z|x_0)}{p_{\theta}(x_0, z)}],
\end{equation} where $q$ and $p_\theta$ are data distribution and learned approximated distribution respectively. $x_0$ is the data sample, $x_{1:T}$ and $z$ are the corresponding latent variables of DDMs and VAEs.
In contrast to the VAE, the primary distinction lies in the representation of latent variables. The VAE incorporates a single latent variable $z$, whereas the DDM introduces multi-step latent variables $x_{1:T}$.
\vspace{-2mm}

\paragraph{Diffusion Phase:} The diffusion process of the DDMs, as a parameterized Markov chain, can be factorized as follows according to the first-order Markov property:
\vspace{-2mm}
\begin{equation}\label{eq:diff}
    q(x_{1:T}|x_0) = \prod_{t-1}^{T}q(x_t|x_{t-1}),    q(x_t|x_{t-1}) = \mathcal{N}(x_t; \sqrt{1-\beta_t}x_{t-1}, \beta_t\mathbf{I}),
\end{equation} where $\mathcal{N}$ is the Normal distribution. At each stage, a small quantity of Gaussian noise is added to corrupt the data, with the variance schedule $\beta_t$ dictating the amount of Gaussian noise introduced at each step. Applying the reparameterization trick, the $x_t^n$ can be formalized as the following closed form with respect to the original input $x_0^n$:
\vspace{-2mm}
\begin{equation}\label{eq:q(x_t|x_0)}
    q(x_t|x_0) = \mathcal{N}(x_t; \sqrt{\bar{\alpha}_t}x_0, (1-\bar{\alpha}_t)\mathbf{I}), \quad \alpha_t = 1-\beta_t, \quad \bar{\alpha}_t = \prod_{i=1}^t \alpha_i.
\end{equation}

\vspace{-2mm}
\paragraph{Denoising Phase:} The primary aim of DDMs is to learn a denoising model capable of reversing the diffusion phase at each step. Consequently, the reverse phase can initially be factorized as follows:
\vspace{-2mm}
\begin{equation}
\begin{split}
        p_\theta(x_0,x_{1:T})& = p(x_T)\prod_{t=1}^T p_\theta(x_{t-1}|x_t), \\
    p_\theta(x_{t-1}|x_t)& = \mathcal{N}(x_{t-1}; \mu_{\theta}(x_t, t), \Sigma_\theta(x_t, t)).
\end{split}
\end{equation} The denoising step is parameterized with learnable $\mu_\theta$ and $\Sigma_\theta$.

\paragraph{Optimization:} In order to optimize the parameter $\theta$, we need to minimize the KL-divergence between $q(x_{t-1}|x_t)$ and $p_{\theta}(x_{t-1}|x_t)$. Given that $q(x_{t-1}|x_t)$ is unknown, the posterior distribution $q(x_{t-1}|x_t, x_0)$ can be computed using $q(x_{t-1}|x_0)$, $q(x_t|x_0)$, and $q(x_t|x_{t-1})$ according to Bayes' rule. The step-wise objective function derived from Eq.~\ref{eq:L_DDM} can be expressed as follows:

\begin{equation}
\begin{split}
        \mathcal{L}_{ELBO} &= \mathbb{E}_q [D_{KL}[q(x_T|x_0)||p(x_T)] \\
        &+ \sum_{t>1}D_{KL}[q(x_{t-1}|x_t, x_0)||p_\theta(x_{t-1}|x_t)] - \log p_\theta(x_0|x_1)],
\end{split}
\end{equation}
where $D_{KL}$ is the KL-divergence.
For the simplified objective, the standard approach involves randomly sampling a fixed number of $t$ and minimizing the second term during each iteration, which can be expressed as follows:
\vspace{-2mm}
\begin{equation}\label{eq:simple}
    L_{simple} = ||\epsilon - \epsilon_\theta(\sqrt{\bar{\alpha}_t}x_0 + \sqrt{1-\bar{\alpha}_t}\epsilon, t)||^2, \quad \alpha_t = 1-\beta_t, \quad \bar{\alpha}_t = \prod_{i=1}^t \alpha_i,
\end{equation} where $\epsilon_\theta$ is the denoiser that removes the Gaussian noise added to $x_0$ at step $t$, and $\epsilon$ denotes the Gaussian Noise.
\vspace{-2mm}
\paragraph{Limitations:} While DDMs have achieved significant success in continuous tasks~\cite{ddpm, dalle2}, there are few triumphs in discrete space, primarily due to the following constraints: 
1), The collapse of the denoising objective. Specifically, each token's representation is randomly initialized (sampled from a normal distribution) and learned during the optimization process. However, in the original DDMs scenario, input features are fixed and distinguishable from each other. Simply applying DDMs to randomly initialized token representations may result in representation collapse~\cite{gao2022difformer}. More concretely, $x_0$ in Eq.~\ref{eq:simple} is also Gaussian noise, rather than an image containing abundant information. As a result, the denoising loss function learns little from the early stages and collapses to a trivial representation solution initially. In our experiments, we observed a similar phenomenon. For the sequential recommendation task, dataset characteristics such as learnable item input features, sparsity, cold-start, and long-tail distribution exacerbate the collapse. 2), The incompatibility of non-autoregressive generation. The autoregressive sequential recommendation is the \textit{de-facto} mechanism. In contrast, DDMs is optimized to reconstruct original image samples in an autoencoder manner. In sequential recommendation, the goal is to predict the next engaged items based on historical interactions. Designing an appropriate diffusion model that retains the merit of high-fidelity generation power necessitates conditional generation rather than generating the target sample from randomly initialized noise. Therefore, we introduce a new optimization framework that leverages the powerful step-wise high-resolution conditional generation capabilities for sequential recommendation.
\vspace{-2mm}
\paragraph{Problem Definition}
In sequential recommendation, we have a set of users $\mathcal{U}$, a set of items $\mathcal{X}$, and a set of user interaction sequences $\mathcal{S} = \{S^1, S^2, \dots, S^{|\mathcal{U}|}\}$. Each sequence comprises a varying number of items $S^i = \{x^1, x^2, \dots, x^{|S^i|}\}$, arranged in chronological order. Sequential recommendation aims to predict the next engaged items $x^{|S^i|}$ for a specific user $u_i$, given their previous interaction sequence $\{x^1, x^2, \dots, x^{|S^i|-1}\}$. In this paper, \modelname generates the representation of items $x^{|S^i|}$ in $T$ steps, conditional on the representation of the historical interaction sequence. For ease of notation, we denote $\mathbf{x^n_{t}}$ as the sample of the target item embedding at diffusion step $t$.

\section{Methodology}

\begin{figure*}
    \centering
    \includegraphics[width=0.9\textwidth]{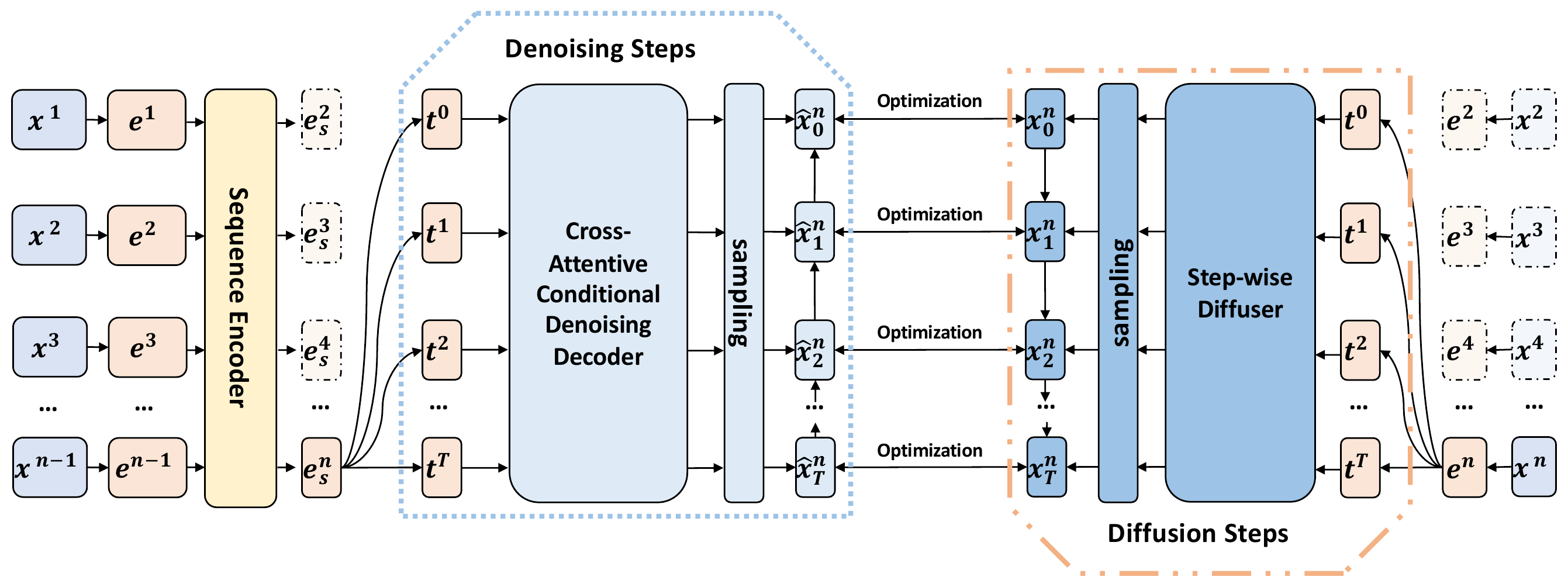}
    \caption{Framework of \modelname. From left to right, we initially employ a sequence encoder to process and encode the sequential patterns derived from prior user interactions. The encoder's output serves as key and value data, combined with the denoising step indicator $t$ as the query for the following cross-attentive conditional denoising decoder. This decoder is utilized to predict the denoised mean of target item embeddings $\mathbf{\hat{x}}^n$ at each denoising step, sampled via the reparameterization trick. Conversely, from right to left, we introduce our step-wise diffuser, responsible for the gradual addition of noise and the creation of corrupted target item embeddings $\mathbf{x^n}$ at each diffusion step, also sampled through the reparameterization trick. Cross-divergence loss and contrastive loss are applied at every step.
    } 
    \vspace{-4mm}
    \label{fig:framework}
\end{figure*}

In this section, we present the comprehensive methodology of \modelname and provide an illustration of the framework in Figure~\ref{fig:framework}. Generally, we first retrieve sequence representations from the item embedding lookup table and feed them into the self-attentive sequence encoder to learn sequence representations based on historical item interactions. These representations act as conditioned information for the cross-attentive conditional denoising decoder. The decoder is aware of the specific step, used as the query of a cross-attentive layer, enabling it to predict the corresponding noise added at each diffusion step. From the item embedding lookup table, we obtain the target sequence embedding, which is shifted one step to the right. We employ the step-wise diffuser to progressively introduce Gaussian Noise at each step. This serves as the target sequence representation for each denoising step.
\vspace{-2mm}

\subsection{Sequence Encoder} 

Previous methods predominantly concentrate on denoising from a randomly initialized Gaussian noise and generating sentences non-autoregressively~\cite{diffusion-lm,gong2022diffuseq, gao2022difformer}. However, in the sequential recommendation, predicting the next item based on historical interaction records autoregressively has proven to be more effective~\cite{SASRec, wang2022contrastvae}. Consequently, in this paper, we employ a self-attentive encoder to learn hidden representations of historical interactions, which can then be fed into the subsequent cross-attentive conditional denoising decoder for the multi-step generation. We also incorporate positional embeddings and attention masks to prevent attending to future interactions. Formally, given a historical sequence ${x^1, x^2, \dots, x^{n-1}}$, we first retrieve item embeddings from the lookup table and construct the input sequence embedding $\mathbf{E} = [\mathbf{e}^1 + \mathbf{p}^1, \mathbf{e}^2+\mathbf{p}^2, \dots, \mathbf{e}^{n-1}+\mathbf{p}^{n-1}]\in \mathbb{R}^{(n-1)\times d}$, where $n = |S^i|$, $\mathbf{e^i}$ and $\mathbf{p^i}$ are the item embedding and position embedding at position $i$. Next, we input this sequence embedding into a self-attention layer (SA):
\begin{equation}
    \mathbf{e_s} = \text{SA}(\mathbf{E}) = \text{Softmax}\left(\frac{(\mathbf{E}\mathbf{W}^Q)(\mathbf{E}\mathbf{W}^K)^{\top}}{\sqrt{d}}\right) (\mathbf{E}\mathbf{W}^V)  \in \mathbb{R}^{(n-1) \times d},
\end{equation} where $\mathbf{W}^Q$, $\mathbf{W}^K$, and $\mathbf{W}^V$ are the learnable parameters, $d$ is the dimension of the embedding, and $\sqrt{d}$ is the normalization factor to avoid large values in the softmax function. 

\subsection{Cross-attentive Conditional Denoising Decoder}

In prior approaches, the denoiser's aim is to learn the reverse denoising process of the associated diffusion step, $p_\theta(x_{t-1}|x_t)\sim \mathcal{N}(\mu_\theta(x_t, t), \hat{\beta_t}\mathbf{I})$, where $\hat{\beta_t} = \frac{1-\bar{\alpha}_{t-1}}{1-\bar{\alpha}_{t}}$ is the closed form of variance of posterior distribution $q(x_{t-1}|x_t, x_0)$, enabling the model denoiser to reconstruct the original input. For sequential recommendation tasks, the objective is to predict subsequent items based on historical interactions in an autoregressive manner. Therefore, rather than generating sequence representations from uncontrollable, randomly initialized Gaussian noise, we opt to integrate the denoiser within the conditional generation framework with a conditional denoising decoder. Drawing inspiration from the performance enhancement strategy of self-conditioning, which entails making the denoiser cognizant of their estimates from the preceding denoising step~\cite{gao2022difformer}—an approach conceptually akin to incorporating residual connections within the denoising chain—we introduce the direct-condition mechanism. This mechanism allows the decoder to directly condition the sequence embeddings from the encoder at every denoising step. Direct conditioning can improve performance by circumventing the long-dependence chain in the denoising phase and directly estimating the noise at each diffusion step from the conditioning sequence embeddings and step indicators. Moreover, this novel design facilitates model convergence to optimal solutions without necessitating large diffusion steps. Consequently, we condition the reverse denoising phase on the preceding sequence representations, expressed as $p_\theta(\mathbf{\hat{x}_{t}}|\mathbf{e_s}, t)\sim \mathcal{N}(\mathbf{\mu_\theta}(\mathbf{e_s}, t), \hat{\beta}_t\mathbf{I})$, where $\mathbf{e_s}$ denotes the encoded historical interactions using the sequence encoder. 

In contrast to earlier methods~\cite{TimeGrad} that maintain only the final position's representation as the sequence representation, we strive to preserve as much information as possible due to the sparse nature of sequential recommendations. Hence, we select a cross-attention architecture as the denoising decoder instantiation capable of taking the entire sequence representation and corresponding step indicator as input. Formally, given a sequence embedding $\mathbf{e_s}$ and the relevant diffusion step $t$, the direct conditional denoising decoder is anticipated to predict the corrupted samples at this diffusion step. Initially, we acquire a learnable embedding $\mathbf{e_t}$ for the indicator $t$ from a time lookup embedding table and expand it to the dimension of $\mathcal{R}^{(n-1)\times d}$, ensuring that every previous hidden embedding is conscious of the same diffusion step. We define the cross-attention (CA) as follows:
\vspace{-2mm}
\begin{equation}
    \mathbf{\mu_\theta^{1:n}}(\mathbf{e_s}, t)=CA(\mathbf{e_s}, \mathbf{e_t}) = \text{Softmax}\left(\frac{(\mathbf{e_t}\mathbf{W}^Q)(\mathbf{e_s}\mathbf{W}^K)^{\top}}{\sqrt{d}}\right) (\mathbf{e_s}\mathbf{W}^V) .
\end{equation} It is worth noting that we mask the items in the future to ensure the autoregressive generation mechanism. Given the predicted mean and the corresponding variance, we can sample the predicted target item embedding at denoising step $t$ using the reparameterization trick:
\begin{equation}\label{eq:p_sample}
    \mathbf{\hat{x}_t^n} = \mathbf{\mu_\theta^n} + \hat{\beta}_t \epsilon, \quad \epsilon\sim \mathcal{N}(0, \mathbf{I}).
\end{equation}
\vspace{-4mm}
\subsection{Step-wise Diffuser}

The goal of the step-wise diffuser is to incrementally introduce Gaussian noise to the original data, creating corrupted data for each step. As per Eq.~\ref{eq:diff}, we predefine a noise schedule $\beta_t$ and corresponding conditional Gaussian distribution $q(\mathbf{x_t^n}|\mathbf{x_{t-1}^n})$ for every diffusion step, where $n$ represents the sequence's final target item. We can employ the reparameterization trick to sample the $\mathbf{x_t^n}$ at any step $t$ using $\mathbf{x_0^n}$, via the closed form Eq.~\ref{eq:q(x_t|x_0)}:
\vspace{-2mm}
\begin{equation}\label{eq:q_sample}
    \mathbf{x_t^n} = \sqrt{\bar{\alpha}_t}\mathbf{x_0^n} + \sqrt{1-\bar{\alpha}_t}\epsilon, \quad \epsilon\sim \mathcal{N}(0,\mathbf{I}),
\end{equation} which is used as the corrupted target to be reconstructed at step $t$.

\subsection{Optimization}

As previously discussed in Sec.~\ref{sec:prelim}, the traditional DDM is designed to reconstruct a sample by removing the Gaussian noise added to it. Consequently, the learning objective is to learn the denoising function $p_\theta(x_{t-1}|x_t)$ for the prior diffusion step, minimizing the KL divergence $D_{KL}[q(x_{t-1}|x_t, x_0)||p_\theta(x_{t-1}|x_t)]$ at each diffusion step. In the inference phase, the conventional DDM begins with Gaussian noise and generates an image by iteratively removing the noise. However, the primary goal of sequential recommendation is to predict the next engaged item based on historical interactions. Firstly, we expect the model to generate high-quality item embeddings conditional on historical interactions, which implies the model should be capable of generating sequence embeddings shifted by one position from the preceding sequence embedding. Secondly, in contrast to image or language generation, the sequential recommendation is a retrieval task requiring the model to effectively rank items, giving higher scores to target items compared to non-interest items. Thirdly, the long-tail and sparse nature of sequential recommendation tasks exacerbate the denoising objective collapse issue in conventional DDMs. Bearing these considerations in mind, we devise a new optimization paradigm that leverages step-wise generation while remaining cognizant of rankings and preventing collapse.

\paragraph{Denoising Diffusion Optimization} To ensure conditional generation, we aim to optimize the KL divergence of the conditional generation function:
\vspace{-2mm}
\begin{equation}\label{eq:KL}
    dis(\mathbf{x_t^n}, \mathbf{\hat{x}_{t}^n}) = D_{KL}[q(\mathbf{x_{t}^n}| \mathbf{x_0^n})||p_\theta(\mathbf{\hat{x}_{t}^n}|\mathbf{e_s}, t)]\propto ||\mathbf{x_t}^n - \mathbf{\hat{x}_t}^n||^2 \propto -\mathbf{x_t^{nT} \hat{x}_t^n}.
\end{equation}
Instead of predicting noise from the target item embedding $\mathbf{x_{t-1}}$ of diffusion step $t-1$, we can directly use the previous sequence embedding to predict the corrupted target item embedding at each diffusion step. This approach not only ensures conditional generation taking advantage of step-wise generation but also avoids accumulating bias during step-wise inference. Since we can directly predict the corrupted sample without knowing $\mathbf{x_{t-1}}$, we can omit this variable in our learning objective. We sample both corrupted target item embeddings and predicted target item embeddings according to Eq.~\ref{eq:q_sample} and Eq.~\ref{eq:p_sample}. In order to account for the model's ranking capability, it is essential not only to minimize the dissimilarity between predicted item embeddings and the target item embeddings but also to maximize the dissimilarity between predicted item embeddings and other irrelevant items. Therefore, rather than merely minimizing the KL-divergence as the objective, we employ the KL-divergence as a generation dissimilarity metric for recommendation purposes.

\paragraph{Cross-Divergence Loss}
Relying solely on the generation objective, the model tends to maximize the similarity between the predicted mean and corrupted target item embedding. However, since all item embeddings are randomly initialized and optimized dynamically, the model may learn trivial item representations where every pair of item embeddings is highly similar, resulting in high-ranking scores for all items. To circumvent this issue, we require the KL divergence between the predicted mean and target item embedding to be smaller than that between the predicted mean and unrelated item embeddings. Consequently, we introduce the cross-divergence loss using Eq.~\ref{eq:KL} as a dissimilarity metric at each denoising step $t$:
\begin{equation}\label{eq:ce}
\begin{split}
        \mathcal{L}_{cd}^t =& \frac{1}{N}\sum_n \log(\sigma(-D_{KL}[q(\mathbf{x_t^n}|\mathbf{x_0^n})||p_\theta(\mathbf{\hat{x}_t^n}|\mathbf{e_s}, t)])) \\
        +&  \log(1-\sigma(-D_{KL}[q(\mathbf{x_t'^n}|\mathbf{x_0'^n})||p_\theta(\mathbf{\hat{x}_t^n}|\mathbf{e_s}, t)]))],
\end{split}
\end{equation} 
where $\mathbf{x_0'^n}$ is the embedding of a randomly sampled negative target item that has never appeared in the user's historical interactions. 

\vspace{-2mm}
\paragraph{Contrastive Loss}
To endow the model with robustness against the noisy interaction, we expect the model to predict consistent item engagement given a certain quantity of noise interpolation. Therefore, inspired by the nature of the diffusion model—which adds Gaussian Noise to target item embeddings and predicts Gaussian Noise with the denoiser—we incorporate a simple yet effective multi-view contrastive learning approach into our framework.

The first view caters to the objective of sequential recommendation. Since we anticipate that the denoised item embedding from the conditional denoising decoder should resemble the target item embedding while distancing itself from other irrelevant items, we optimize the in-view InfoNCE loss as follows:
\begin{equation}\label{eq:in-view}
    \mathcal{L}_{in}^t = \frac{1}{N}\sum\limits_{i=1}^{N}  \log \frac{\exp(\hat{\mathbf{x}}_{t}^{i\top} \mathbf{x}_t^i / \tau)}{\sum\limits_{j} \exp(\hat{\mathbf{x}}^{i\top}_t \mathbf{x}_t^j / \tau) + \sum\limits_{j} \mathbbm{1}_{[j\neq i]}\exp(\hat{\mathbf{x}}^{i\top}_t \mathbf{\hat{x}}_t^j / \tau)},
\end{equation} 
where $\mathbf{\hat{x}_t^i}$ is the output of conditional denoising decoder of position $i$ at denoising step $t$, $\mathbf{x_t^i}$ is the output of step-wise diffuser at diffusion step $t$ at position $i$ of the sequence.

From another perspective, we expect the model to be robust to noise in the input data. In other words, when a small interpolation is introduced to the input sequence, the conditional denoiser should still predict similar target item sequence representations. As a result, we initially acquire the augmented view by modifying the input sequence through random cropping, shuffling, and masking. Then, we incorporate a supplementary cross-view InfoNCE loss, which aims to maximize the agreement between the original view and its augmented counterpart, as detailed below:
\begin{equation}\label{eq:cross-view}
    \mathcal{L}_{cross}^t = \frac{1}{N}\sum\limits_{i=1}^N  \log \frac{\exp(\hat{\mathbf{x}}_{t}^{i\top} \Tilde{\mathbf{x}}_{t}^i / \tau)}{\sum\limits_{j} \exp(\hat{\mathbf{x}}^{i\top}_t  \Tilde{\mathbf{x}}_{t}^j / \tau) + \sum\limits_{j} \mathbbm{1}_{[j\neq i]}\exp(\hat{\mathbf{x}}^{i\top}_t \mathbf{\hat{x}}_t^j / \tau)},
\end{equation}
where $ \Tilde{\mathbf{x}}_{t}^i$ is the output of conditional denoising decoder at diffusion step $t$, position $i$ of augmented view.

Since the noise added to the target item embedding increases as the diffusion step progresses, more information is lost at higher diffusion steps. Intuitively, to avoid focusing too much on reconstructing non-informative noise, we rescale the loss term of each diffusion step by dividing it by the step indicator. Furthermore, unlike previous methods that randomly sample step indicators for optimization, we explicitly calculate the loss term for every diffusion step. The final optimization objective can be formalized as follows:
\begin{equation}\label{eq:rescale}
    \mathcal{L}^{re} = \sum_{t=0}^{T}\frac{1}{t+1}(\mathcal{L}_{cd}^t+ \lambda (\mathcal{L}_{in}^t + \mathcal{L}_{cross}^t)).
\end{equation}

\section{Experiments}
In this section, we evaluate the proposed \modelname on sequential recommendation tasks to examine the following research questions:
\begin{itemize}
    \item \textbf{RQ1}: How does \modelname perform compared to state-of-the-art sequential recommendation models.
    \item \textbf{RQ2}: What are the contributions of each key designs of \modelname.
    \item \textbf{RQ3}: How does \modelname perform with respect to different subsets of sequences and different denoising steps.
\end{itemize}

\paragraph{Dataset}
In this paper, we conduct experiments on four Amazon datasets~\cite{amazon}: \textit{Office}, \textit{Beauty}, \textit{Tools and Home}, and \textit{Toys and Games}. We first filter out users and items with fewer than five interaction records. Next, we sort user interactions with items in chronological order. In line with common practice~\cite{SASRec, S3Rec, Bert4Rec}, for each user, we treat the last engaged item as the test item, the penultimate one as the validation item, and all previous items, excluding these two, as training items. In other words, we split the train, validation, and test datasets using all previous interactions except for the last two, second-to-last interactions, and the last ones from all users, respectively. We report the statistics of the datasets in Table~\ref{tab:data_stat}.
\vspace{-2mm}
\begin{table}[H]
\centering
\caption{Statistics of datasets, we report the number of users, number of items, number of interactions, number of interactions per item, and the averaged sequence length.}
\vspace{-2mm}
\label{tab:data_stat}
\small
\begin{tabular}{@{}c|ccccc@{}}
\hline
Dataset& \#Users &\#Items &\#Interactions &\#Ints / item & Avg. seq. len.\\
\hline
Beauty & 22,363 & 12,101 & 198,502 & 16.40 & 8.3 \\
Toys & 19,412 & 11,924 & 167,597 & 14.06 & 8.6 \\
Tools & 16,638 & 10,217 & 134,476 & 13.16 & 8.1 \\
Office & 4,905 & 2,420 & 53,258 & 22.00 & 10.8 \\
\hline
\end{tabular}
\end{table}
\vspace{-2mm}
\paragraph{Baseline Models} We conduct the overall comparison with these three related types of state-of-the-art methods:
\begin{itemize}
    \item Generative Models: \textbf{MultVAE}~\cite{MultVAE}, \textbf{ACVAE}~\cite{ACVAE}, \textbf{ContrastVAE}~\cite{wang2022contrastvae}. MultVAE models the interactions of a user as multinomial distribution and optimizes the VAE by reconstructing the interactions. ACVAE introduces the concept of adversarial variational Bayes and mutual information maximization to optimize the VAE for the sequential recommendation. ContrastVAE introduces the objective named ContrastELBO, a variational augmentation strategy to optimize the VAE via maximizing the mutual information among latent variables and the next item embedding reconstruction likelihood. 
    \item Contrastive Models: \textbf{CL4Rec}~\cite{CL4Rec}, \textbf{DuoRec}~\cite{DuoRec}, \textbf{CBiT}~\cite{CBiT}. CL4Rec introduces the data augmentation strategies: mask, random shuffle, and random crop into the sequential recommendation task and optimizes the model via InfoNCE~\cite{infonce} loss. DuoRec further improves the contrastive learning paradigm in sequential recommendation via semantic augmentation, which regards the sequences with the same target items as the positive views instead of sequences constructed from data-level augmentations. CBiT further improves Bert4Rec by introducing the additional InfoNCE objective. 
    \item Encoder Models: \textbf{GRU4Rec}~\cite{GRU4Rec}, \textbf{SASRec}~\cite{SASRec}, \textbf{FMLP}~\cite{FMLP}. GRU4Rec first attempts the Recurrent Neural Network for the sequential recommendation, while SASRec first employs a transformer-based encoder to learn sequence and item representation. FMLP further replaces the multi-head self-attention layer of SASRec with the denoising Fourier layer, thus designing a novel encoder only with multi-layer perceptions.
\end{itemize}

\paragraph{Metrics}
To evaluate the performance of our model, we employ ranking-related evaluation metrics, including Recall@N, NDCG@N, and MRR, following common practice~\cite{SASRec, CL4Rec, wang2022contrastvae}. We first obtain the sequence representation using the user interaction records except for the last one. Then, we calculate the ranking score using the dot product between the sequence representation and candidate items (all items in the dataset are used as candidates) and sort these scores in descending order. These sorted ranking scores are used to evaluate Recall, NDCG, and MRR with respect to the last test items.

\paragraph{Implementation Details}

We implement our model using Pytorch and conduct our experiments on an Nvidia V100 GPU with 40G memory. We choose Adam~\cite{adam} as our optimization strategy and use early stopping with patience of 50 to prevent overfitting. We employ a learning rate of $0.001$, batch size of $128$, dropout probability of $0.2$, embedding dimension of $128$, and maximum sequence length of $20$ as our hyperparameters. For a fair comparison, we conduct 30 experiments for every baseline method by randomly selecting hyperparameters such as learning rate, dropout probability, hidden size, number of attention heads, etc.

\subsection{Overall Experiments}

In this paper, we conduct a comprehensive comparison between \modelname and state-of-the-art models, reporting the numerical results in Table~\ref{tab:overall_comp}. Our model \modelname consistently outperforms other models on these four datasets, demonstrating the effectiveness of our model. Specifically, in terms of Recall@1, our model shows substantial improvement with gains of $20.98\%$, $16.67\%$, $17.59\%$, and $18.42\%$ compared to the second-best models on Office, Beauty, Tools, and Toys, respectively. We attribute these improvements to the high-quality item embeddings generated by \modelname. As a result, the items are more distinguishable from negative items and closer to positive targets, increasing the likelihood of ranking the target item first. Concretely, if predicted ranking scores are over-smoothed, i.e., the top-rated scores are more similar to each other, it is more likely to hit the target items when considering the top 10 ranked items. This explains why \modelname has relatively smaller improvements on Recall@10 compared to other metrics. From another perspective, metrics like NDCG@5, NDCG@10, and MRR, which take the ranking position of target items into account, still show impressive improvements. This indicates that our model \modelname ranks target items relatively higher than other models.
\vspace{-2mm}
\begin{table*}[t]
    \caption{Overall Comparison. The best is bolded, and the runner-up is underlined.}
    \vspace{-2mm}
    \label{tab:overall_comp}
    \small
    \centering
    \setlength{\tabcolsep}{0.8mm}{
    \begin{tabular}{l|l|cccccccccccc}
         \hline
         Dataset & Metric & SVAE & ACVAE & ContrastVAE & CL4Rec & DuoRec & CBiT & GRU4Rec & SASRec & Bert4Rec & FMLP &  \modelname & Imp\\
         \hline
   
        \multirow{7}{*}{Office} % updated with the new results beta = 0.04
        & R@1  & 0.0088 & 0.0139 & 0.0194 & 0.0094 & 0.0120 & 0.0198 & 0.0051 & 0.0198 & 0.0137 & \underline{0.0224} & \textbf{0.0271} & 20.98\% \\
        & R@5  & 0.0316 & 0.0457 & 0.0642 & 0.0294 & 0.0330 & 0.0593 & 0.0241 & \underline{0.0656} & 0.0485 & 0.0593 & \textbf{0.0765} & 16.62\%  \\
        & R@10 & 0.0597 & 0.0742 & \underline{0.1052} & 0.0430 & 0.0559 & 0.0917 & 0.0510 & 0.0989 & 0.0848 & 0.0901 & \textbf{0.1091} & 3.71\%  \\
        & N@5  & 0.0202 & 0.0300 & 0.0411 & 0.0194 & 0.0223 & 0.0396 & 0.0149 & \underline{0.0428} & 0.0309 & 0.0414 & \textbf{0.0521}  & 21.73\%  \\
        & N@10 & 0.0292 & 0.0392 & \underline{0.0544} & 0.0237 & 0.0296 & 0.0500 & 0.0234 & 0.0534 & 0.0426 & 0.0513 & \textbf{0.0627} & 15.26\% \\
        & MRR  & 0.0249 & 0.0351 & \underline{0.0463} & 0.0207 & 0.0264 & 0.0437 & 0.0204 & 0.0457 & 0.0408 & 0.0455 & \textbf{0.0548} & 18.36\% \\
        \hline

        \multirow{4}{*}{Beauty} %updated  with T=30
        & R@1  & 0.0014 & 0.0167 & 0.0161 & 0.0045 & 0.0107 & \underline{0.0174} & 0.0079 & 0.0129 & 0.0119 & 0.0154 & \textbf{0.0203} & 16.67\%\\
        & R@5  & 0.0068 & 0.0428 & 0.0491 & 0.0160 & 0.0278 & \underline{0.0512} & 0.0266 & 0.0416 & 0.0396 & 0.0433 & \textbf{0.0542} & 5.86\% \\
        & R@10 & 0.0127 & 0.0606 & 0.0741 & 0.0250 & 0.0403 & \underline{0.0762} & 0.0421 & 0.0633 & 0.0595 & 0.0627 & \textbf{0.0770} & 1.05\%\\
        & N@5  & 0.0041 & 0.0299 & 0.0327 & 0.0103 & 0.0193 & \underline{0.0343} & 0.0172 & 0.0274 & 0.0257 & 0.0297 & \textbf{0.0376} & 9.62\%\\
        & N@10 & 0.0060 & 0.0356 & 0.0407 & 0.0131 & 0.0233 & \underline{0.0424} & 0.0222 & 0.0343 & 0.0321 & 0.0360 & \textbf{0.0447} & 5.42\%\\
        & MRR  & 0.0046 & 0.0310 & 0.0345 & 0.0111 & 0.0201 & \underline{0.0359} & 0.0191 & 0.0291 & 0.0294 & 0.0305 & \textbf{0.0387} & 7.80\% \\
     
         \hline

        \multirow{4}{*}{Tools} % updated with the new results beta = 0.08
        & R@1  & 0.0055 & 0.0090 & \underline{0.0108} & 0.0060 & 0.0058 & 0.0066 & 0.0047 & 0.0103 & 0.0059 & 0.0089 & \textbf{0.0127} & 17.59\%\\
        & R@5  & 0.0118 & 0.0242 & \underline{0.0315} & 0.0189 & 0.0182 & 0.0214 & 0.0154 & 0.0284 & 0.0189 & 0.0251 & \textbf{0.0359} & 13.97\% \\
        & R@10 & 0.0204 & 0.0364 & \underline{0.0483} & 0.0293 & 0.0361 & 0.0347 & 0.0242 & 0.0427 & 0.0319 & 0.0359 & \textbf{0.0522} & 8.07\%\\
        & N@5  & 0.0086 & 0.0166 & \underline{0.0212} & 0.0123 & 0.0120 & 0.0139 & 0.0102 & 0.0194 & 0.0123 & 0.0170 & \textbf{0.0244} & 15.09\%\\
        & N@10 & 0.0114 & 0.0206 & \underline{0.0266} & 0.0156 & 0.0148 & 0.0182 & 0.0129 & 0.0240 & 0.0165 & 0.0204 & \textbf{0.0297} & 11.65\%\\
        & MRR  & 0.0098 & 0.0178 & \underline{0.0227} & 0.0132 & 0.0128 & 0.0154 & 0.0113 & 0.0207 & 0.0160 & 0.0174 & \textbf{0.0253} & 11.45\%\\
     
         \hline

        \multirow{4}{*}{Toys} % updated with the new results T=25
        & R@1  & 0.0022 & 0.0156 & \underline{0.0228} & 0.0067 & 0.0099 & 0.0195 & 0.0066 & 0.0193 & 0.0110 & 0.0189 & \textbf{0.0270} & 18.42\%\\
        & R@5  & 0.0057 & 0.0349 & \underline{0.0591} & 0.0180 & 0.0258 & 0.0525 & 0.0226 & 0.0551 & 0.0300 & 0.0516 & \textbf{0.0665} & 12.52\%\\
        & R@10 & 0.0098 & 0.0492 & \underline{0.0823} & 0.0259 & 0.0360 & 0.0747 & 0.0363 & 0.0797 & 0.0466 & 0.0674 & \textbf{0.0935} & 13.61\%\\
        & N@5  & 0.0038 & 0.0255 & \underline{0.0414} & 0.0124 & 0.0179 & 0.0364 & 0.0148 & 0.0377 & 0.0206 & 0.0357 & \textbf{0.0472} & 14.01\%\\
        & N@10 & 0.0038 & 0.0301 & \underline{0.0489} & 0.0149 & 0.0212 & 0.0435 & 0.0192 & 0.0456 & 0.0260 & 0.0408 & \textbf{0.0559} & 14.31\%\\
        & MRR  & 0.0044 & 0.0270 & \underline{0.0422} & 0.0132 & 0.0182 & 0.0373 & 0.0165 & 0.0385 & 0.0244 & 0.0347 & \textbf{0.0479} & 13.51\%\\
     
         \hline

    \end{tabular}
    }
\end{table*}
\vspace{-2mm}

\subsection{Ablation Study}
\begin{figure*}[h]
    \centering
    \begin{subfigure}{0.24\textwidth}
        \centering
        \includegraphics[width=\linewidth]{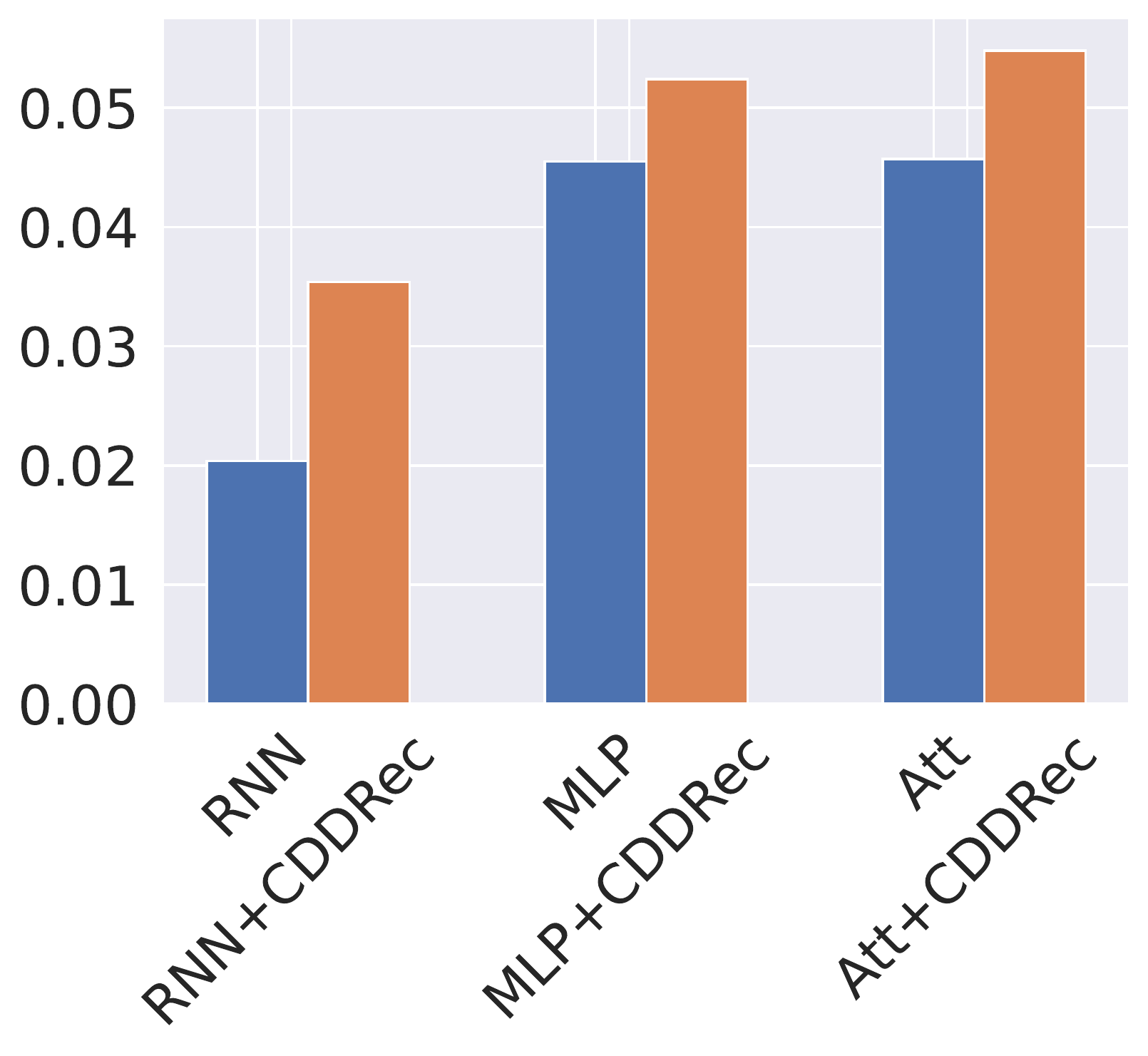}
        \caption{Office}
        \label{fig:subfigure-a}
    \end{subfigure}
    \hfill
    \begin{subfigure}{0.24\textwidth}
        \centering
        \includegraphics[width=\linewidth]{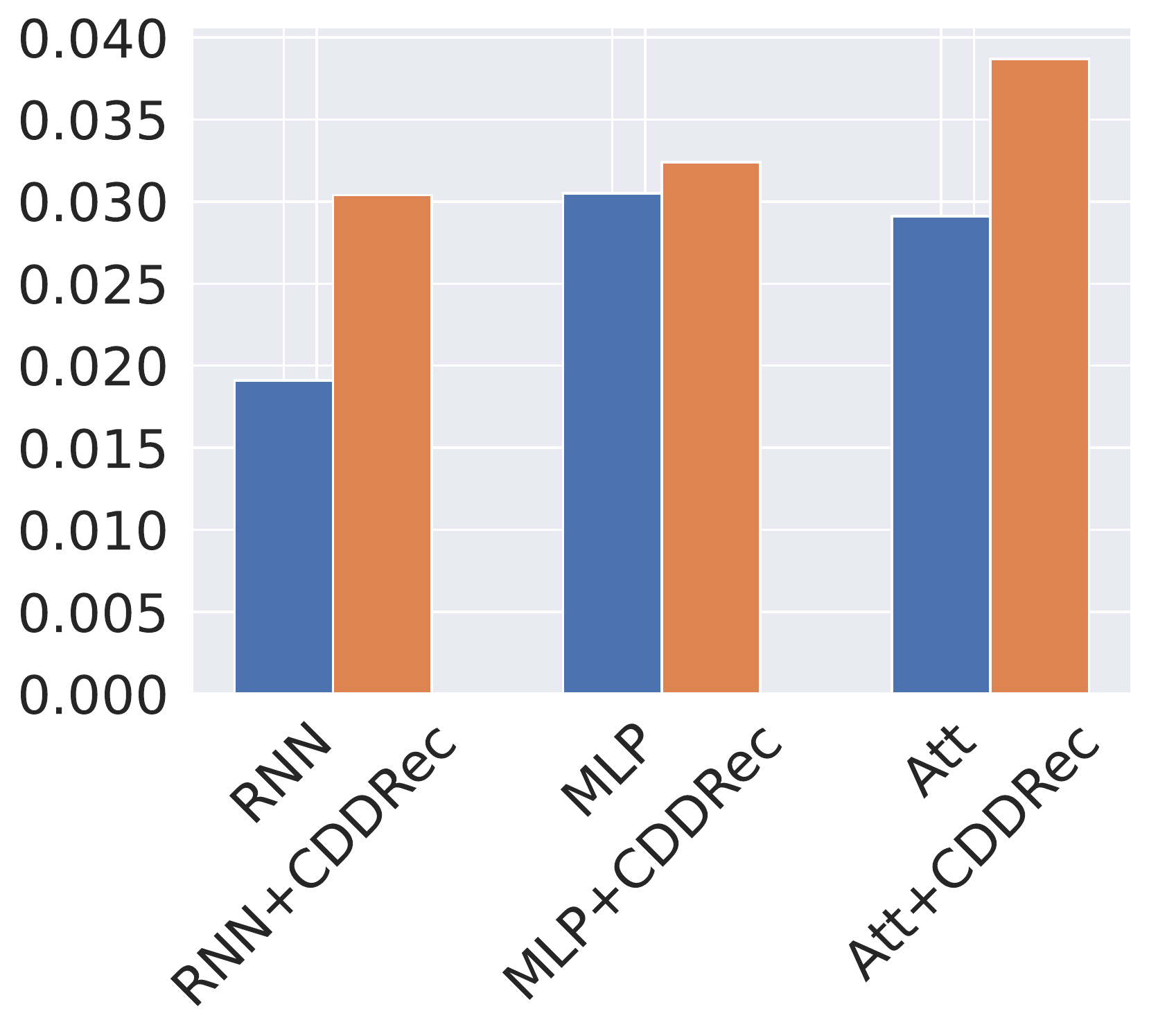}
        \caption{Beauty}
        \label{fig:subfigure-b}
    \end{subfigure}
    \hfill
    \begin{subfigure}{0.24\textwidth}
        \centering
        \includegraphics[width=\linewidth]{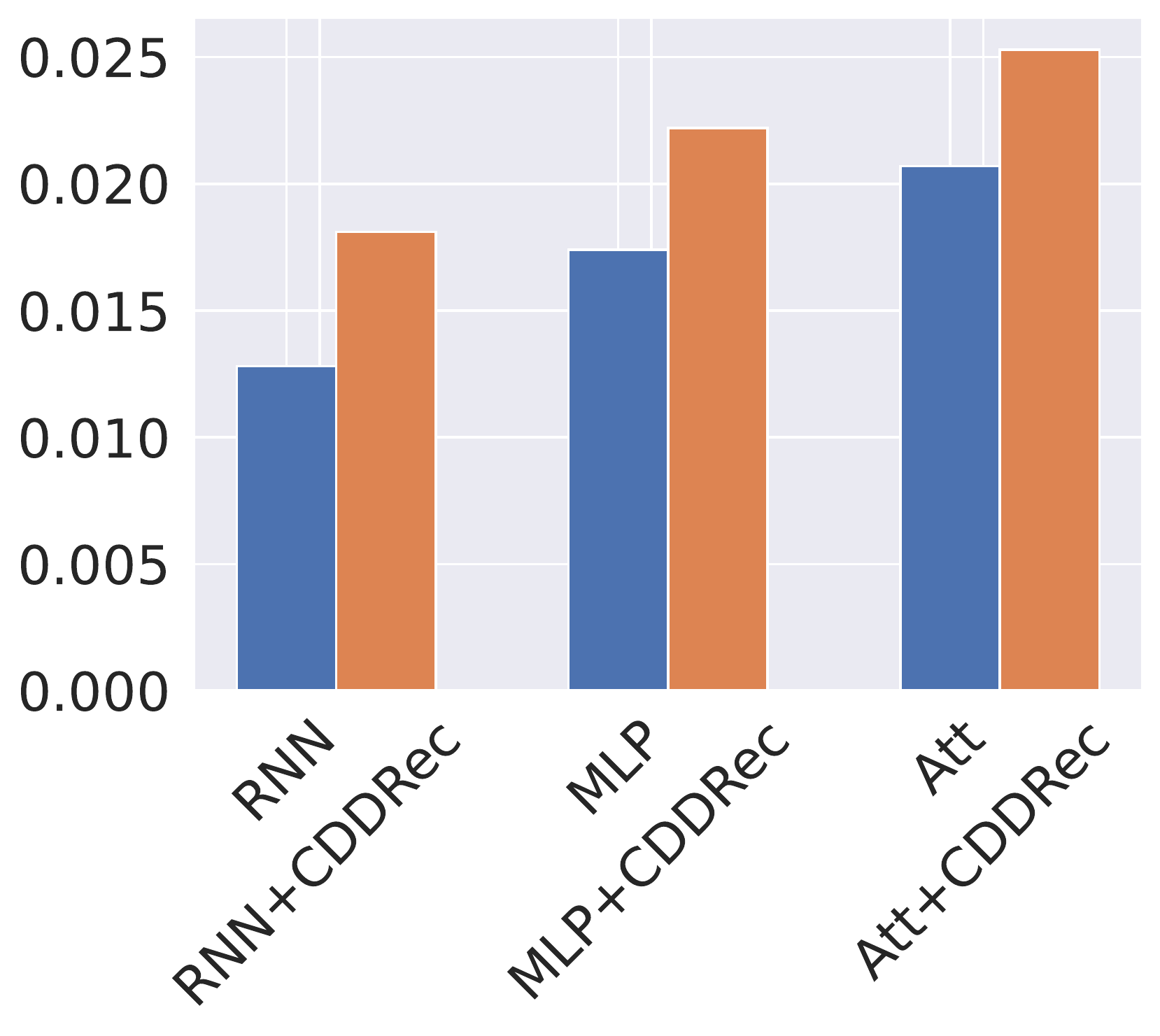}
        \caption{Tool}
        \label{fig:subfigure-c}
    \end{subfigure}
    \hfill
    \begin{subfigure}{0.24\textwidth}
        \centering
        \includegraphics[width=\linewidth]{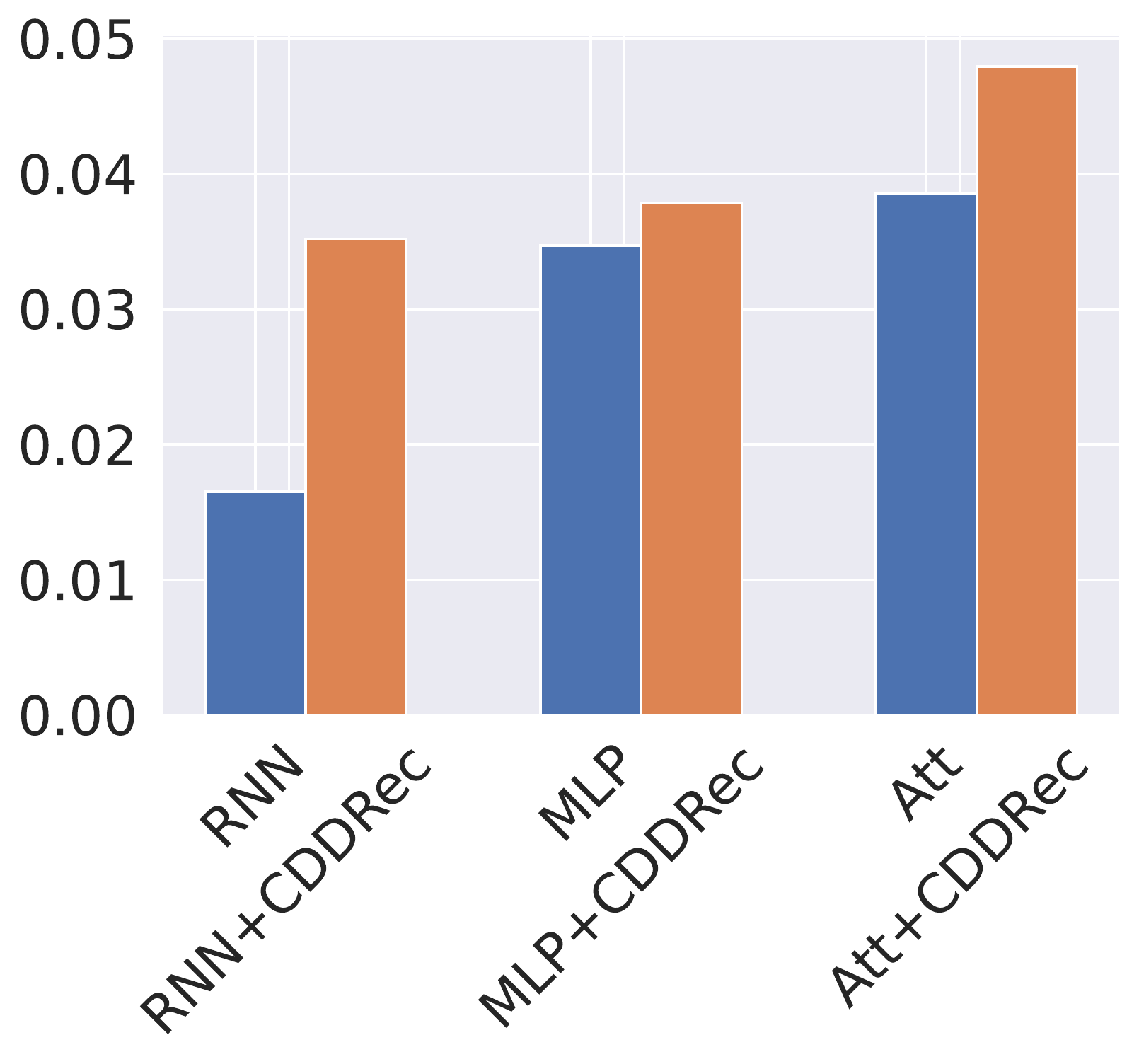}
        \caption{Toy}
        \label{fig:subfigure-d}
    \end{subfigure}
    \vspace{-2mm}
    \caption{The evaluation of \modelname on MRR through four datasets with different sequence encoders. RNN, MLP, and Att are the encoders of GRU4Rec, FMLP, and SASRec respectively.}

    \label{fig:encoders}
\end{figure*}

In this section, we conduct experiments to examine the effectiveness and contributions of each component, rescaled cross-divergence loss, and contrastive loss. We report the experimental results on the Office dataset in Table~\ref{tab:exp-ablation}, Table~\ref{tab:diff_denoise}, and Fig.~\ref{fig:encoders}. From the experiments, we have the following observations: 
\paragraph{Effect of Cross-Divergence loss based on Dissimilar Metric}

In Table~\ref{tab:exp-ablation}, the significant performance drop of $88.88\%$ and $91.42\%$ was observed in columns MSE from single-view and multi-view. MSE loss (Eq.~\ref{eq:KL}) that only minimizes the KL divergence between the predicted item and the target item embeddings can lead to the collapse issues. Furthermore, the multi-view without any regularization exacerbates such collapse issues. In contrast, the cross-divergence loss dramatically improves the performance when introduced as shown in columns $\mathcal{L}_{cd}^{re}$, as it incorporates ranking capability into the diffusion model by maximizing the divergence between predicted item embeddings and randomly sampled negative item embeddings. The comparison between columns $\mathcal{L}_{ce}^{re}$ from single-view and multi-view shows that data augmentation has a positive effect on performance when cross-divergence loss is employed.

\paragraph{Effect of loss term rescale}
By comparing column $\mathcal{L}_{cd}$ and $\mathcal{L}_{cd}^{re}$ in Table~\ref{tab:exp-ablation}, we can conclude that the diffusion-step-rescaled loss term has a positive effect on the overall performance. Due to space limitations, we only report the variants with the adaptive loss for multi-view settings. It is worth noting that, during the experiments, we also observed a consistent performance drop for each variant when using the mean of loss instead of the diffusion-step-rescaled loss.

\paragraph{Effect of contrastive loss}
By comparing columns $\mathcal{L}_{cd}^{re}$+In and $\mathcal{L}_{cd}^{re}$+Cross with the overall performance of \modelname in Table~\ref{tab:exp-ablation}, we can conclude that both the cross-view loss and in-view loss have a positive effect on the overall performance. Furthermore, we can also observe that the in-view contrastive loss has more effect on the overall performance compared to the cross-view loss. Without the In-view InfoNCE loss, the Cross-view InfoNCE loss has negative results, the possible reason is that the cross-view InfoNCE loss only requires the predicted item embeddings to resemble given a certain amount of noise interpolation, ignoring the ranking requirement of models. This phenomenon indicates the necessity of the combination of both in-view and cross-view optimization.
\vspace{-2mm}
\begin{table*}
    \caption{Ablation Study: Single-View indicates the variants that do not have the positive view of the sequences via data augmentation, Multi-View is another way around. MSE implies that we do not apply cross-divergence loss, instead, we optimize the model via Mean squared error loss between predicted item embedding and target item embedding. $\mathcal{L}_{cd}$ is the variant of $\mathcal{L}_{cd}^{re}$ that calculate mean error of all denoising step. "Cross" denotes the cross-view contrastive loss while "In" demonstrates the in-view contrastive loss. "avg.drop" is the average performance drop compared to \modelname.}
    \vspace{-2mm}
    \label{tab:exp-ablation}
    \small
    \centering
    \begin{threeparttable}
        \setlength{\tabcolsep}{2mm}
    {
        \begin{tabular}{l|l|ccc|cccc|cccc}
         \toprule
         \multirow{2}{*}{}
         &Variants&\multicolumn{3}{c|}{Single-View} &\multicolumn{4}{c|}{Multi-View} &\modelname \\
         \hline
         & Metric & MSE & $\mathcal{L}_{cd}$ &  $\mathcal{L}_{cd}^{re}$ & MSE & $\mathcal{L}_{cd}^{re}$ & $\mathcal{L}_{cd}^{re}$+In & $\mathcal{L}_{cd}^{re}$+Cross & \\
         \midrule
         \multirow{7}{*}{Office}
         & R@1  & 0.0016 & 0.0198 & 0.0208 & 0.0006 & 0.0226 & 0.0241 & 0.0218 & \textbf{0.0271} \\
         & R@5  & 0.0096 & 0.0622 & 0.0650 & 0.0075 & 0.0693 & 0.0691 & 0.0693 & \textbf{0.0765} \\
         & R@10 & 0.0153 & 0.0952 & 0.1011 & 0.0133 & 0.1078 & 0.1068 & 0.1070 & \textbf{0.1091} \\
         & N@5  & 0.0055 & 0.0411 & 0.0425 & 0.0041 & 0.0462 & 0.0470 & 0.0460 & \textbf{0.0521} \\
         & N@10 & 0.0073 & 0.0517 & 0.0541 & 0.0060 & 0.0585 & 0.0590 & 0.0581 & \textbf{0.0627} \\
         & MRR  & 0.0066 & 0.0440 & 0.0463 & 0.0054 & 0.0499 & 0.0506 & 0.0500 & \textbf{0.0548} \\
         \midrule
         & avg. drop  & 88.88\% & 19.46\% & 15.54\% &91.42\% & 9.03\% & 7.70\% & 9.78\% & 0\%\\
         \bottomrule
        \end{tabular}
    }
    \end{threeparttable}
\end{table*}

\paragraph{Study of Diffusion and Denoising}

During the experiments, our proposed model has been endowed with denoising capabilities as delineated in Eq.~\ref{eq:p_sample}. Notably, the model is designed to predict the mean of the target item embedding. By employing this sampling procedure, the model can simultaneously predict the noise introduced into the target item embeddings. Accurate prediction of the mean for these perturbed items, devoid of noise, confers the denoising ability upon the model. On the other hand, the diffusion step also follows a sampling process as described by Eq.~\ref{eq:q_sample}. In the absence of this diffusion step, the model would encounter identical original target item embeddings throughout each denoising step.

To scrutinize the impact of these dual processes, we substitute the two sampling steps with the predicted mean and the original item embedding, respectively, and present the experimental findings in Table~\ref{tab:diff_denoise}. The results yield several key observations: Firstly, the diffusion and denoising processes generally contribute positively to the overall performance across all datasets, as evidenced by the performance decline in comparison to our model \modelname. Moreover, the significance of diffusion and denoising varies among datasets. Specifically, the denoising process demonstrates greater importance on the Office and Toys datasets, while the diffusion phase is more crucial for the Beauty and Tools datasets.

An explanation for this variation could be attributed to the differences in sequence length across the datasets. Office and Toys datasets exhibit relatively longer sequences, which could result in a higher likelihood of noisy interactions, thereby rendering the sampling process on the target sequence less effective. Conversely, when dealing with shorter sequences, the diffusion phase that introduces noise may serve as an augmentation strategy, bolstering the model's robustness to noisy interactions.

\vspace{-2mm}
\begin{table*}
    \caption{Study of Diffusion and Denoising. The terms "-Diffusion" and "-Denoising" imply that we substitute diffusion sampling with target item embeddings and denoising sampling with the predicted mean, respectively. "avg.drop" is the average performance drop compared to \modelname.}
    \vspace{-2mm}
    \label{tab:diff_denoise}
    \small
    \centering
    \begin{threeparttable}
        \setlength{\tabcolsep}{2mm}
    {
        \begin{tabular}{l|ccc|ccc|ccc|cc}
         \toprule
         \multirow{2}{*}{}
         &\multicolumn{3}{c|}{-Diffusion} &\multicolumn{3}{c|}{-Denoising}  &\multicolumn{2}{c}{\modelname} \\
         \hline
         & R@1 & MRR & avg.drop & R@1 & MRR & avg.drop &  R@1 & MRR \\
         \midrule
      
         Office & 0.0236 & 0.0496 & 11.20\% & 0.0222 & 0.0490 & 14.33\%  & \textbf{0.0271} &\textbf{ 0.0548}\\ % already updated
         Beauty & 0.0154 & 0.0342 & 17.88\% & 0.0179 & 0.0365 & 8.75\% &   \textbf{0.0203} & \textbf{0.0387} \\ % updated all
         Tools  & 0.0112 & 0.0226 & 11.24\% & 0.0116 & 0.0243 & 6.31\% & \textbf{0.0127} &\textbf{0.0253} \\ % updated all
         Toys   & 0.0267 & 0.0464 & 2.12\% & 0.0264 & 0.0462 & 2.89\%  &\textbf{0.0270} & \textbf{0.0479} \\ % updated all
         \bottomrule
        \end{tabular}
    }
    \end{threeparttable}
\end{table*}

\paragraph{Study of Conditional Encoder}

In this study, we introduce a conditional denoising decoder and an associated optimization approach. We posit that incorporating the conditional denoising decoder into existing encoders and employing our optimization schema can enhance the performance of these methods. To substantiate our claim, we conduct experiments wherein we integrate our conditional denoising decoder with the encoders of GRU4Rec, FMLPRec, and SASRec, in conjunction with our optimization technique. The outcomes of these experiments are presented in Figure~\ref{fig:encoders}. Our findings reveal a consistent performance improvement across all encoders and datasets, suggesting that the proposed framework is not only compatible with existing encoder-based methodologies but also contributes positively to their original performance levels.
\vspace{-2mm}

\subsection{Hyperparameter Sensitivity}
\begin{figure*}[h]
    \centering
    \begin{subfigure}{0.24\textwidth}
        \centering
        \includegraphics[width=\linewidth]{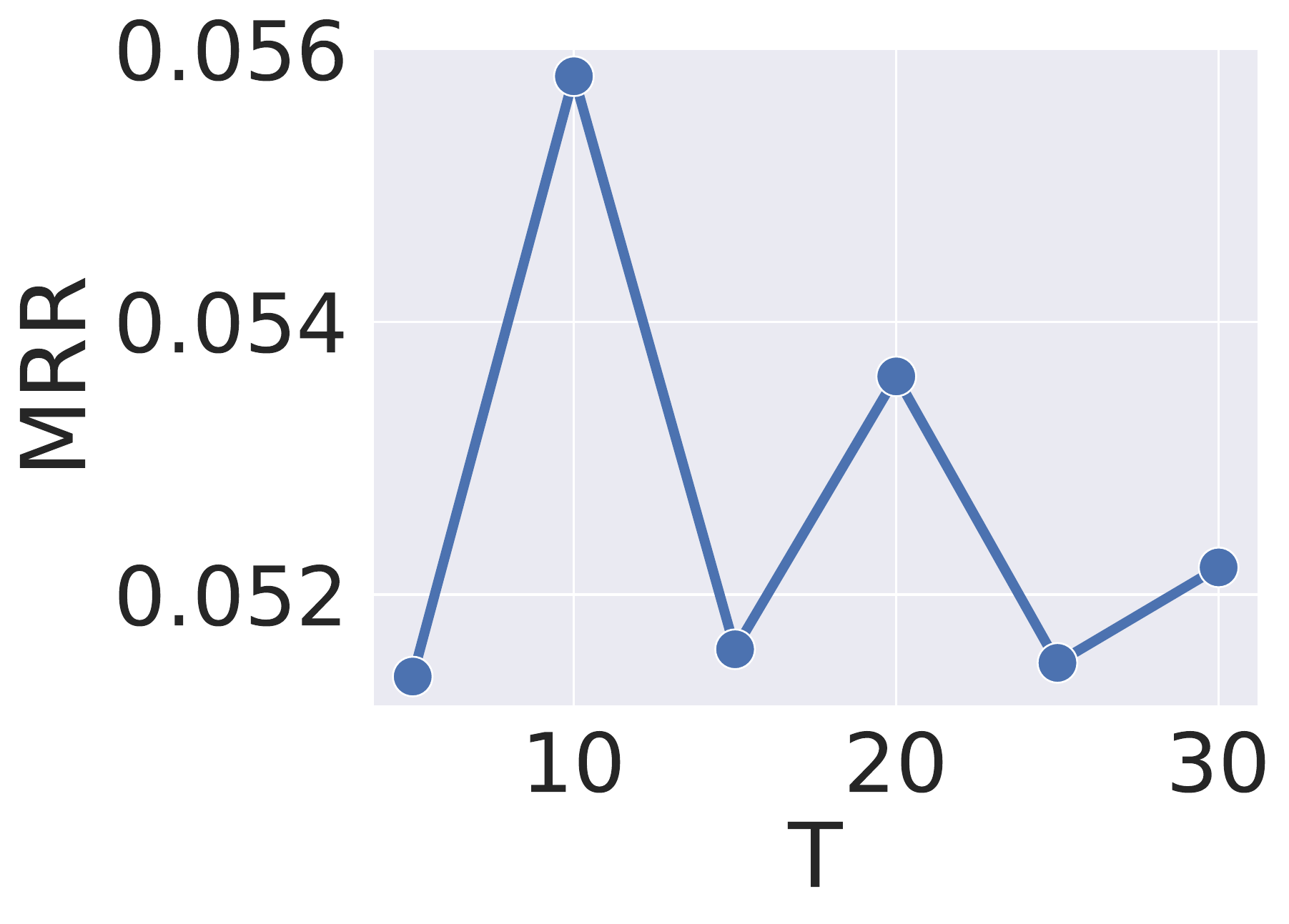}
        \caption{Office}
        \label{fig:subfigure-a}
    \end{subfigure}
    \hfill
    \begin{subfigure}{0.24\textwidth}
        \centering
        \includegraphics[width=\linewidth]{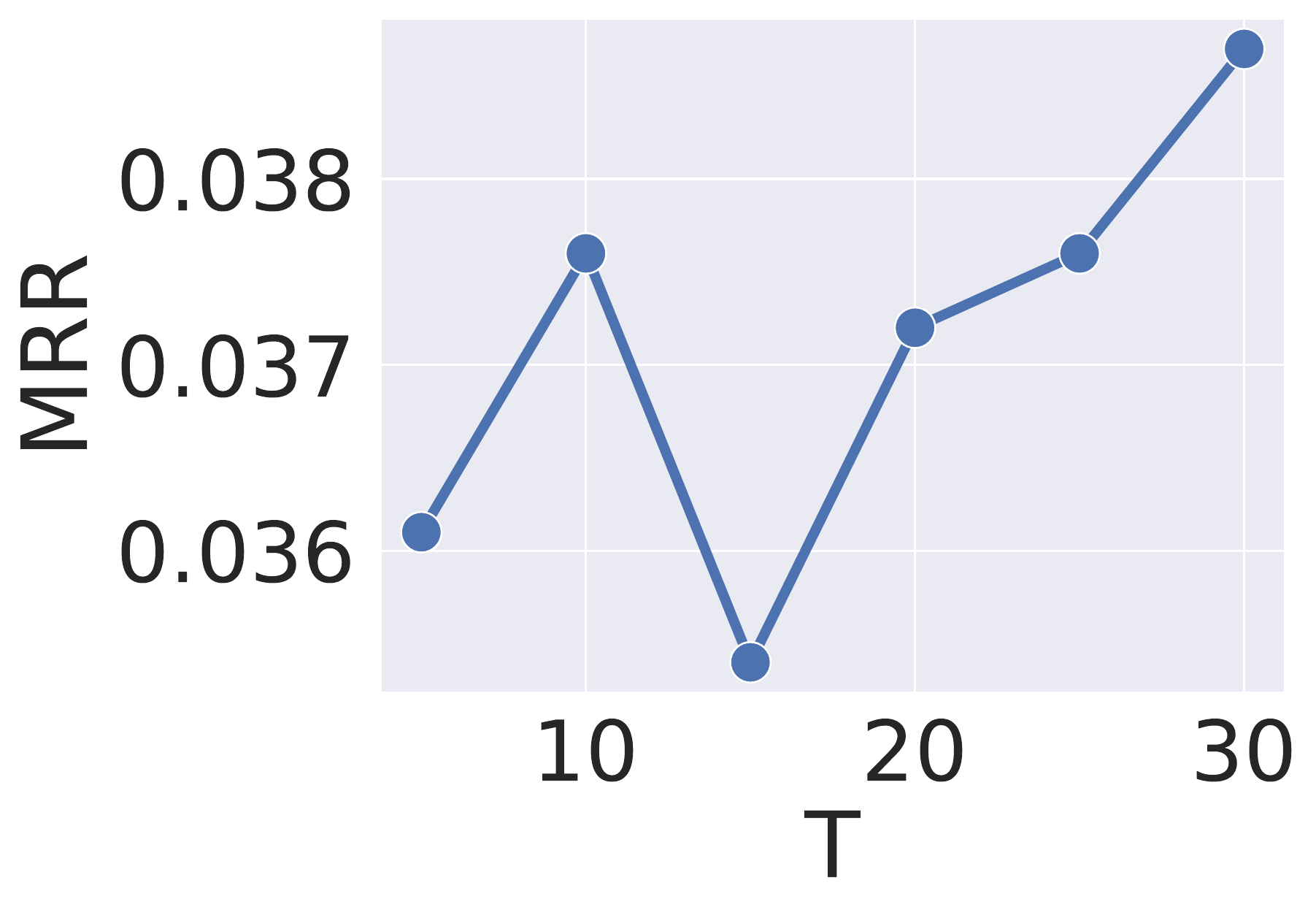}
        \caption{Beauty}
        \label{fig:subfigure-b}
    \end{subfigure}
    \hfill
    \begin{subfigure}{0.24\textwidth}
        \centering
        \includegraphics[width=\linewidth]{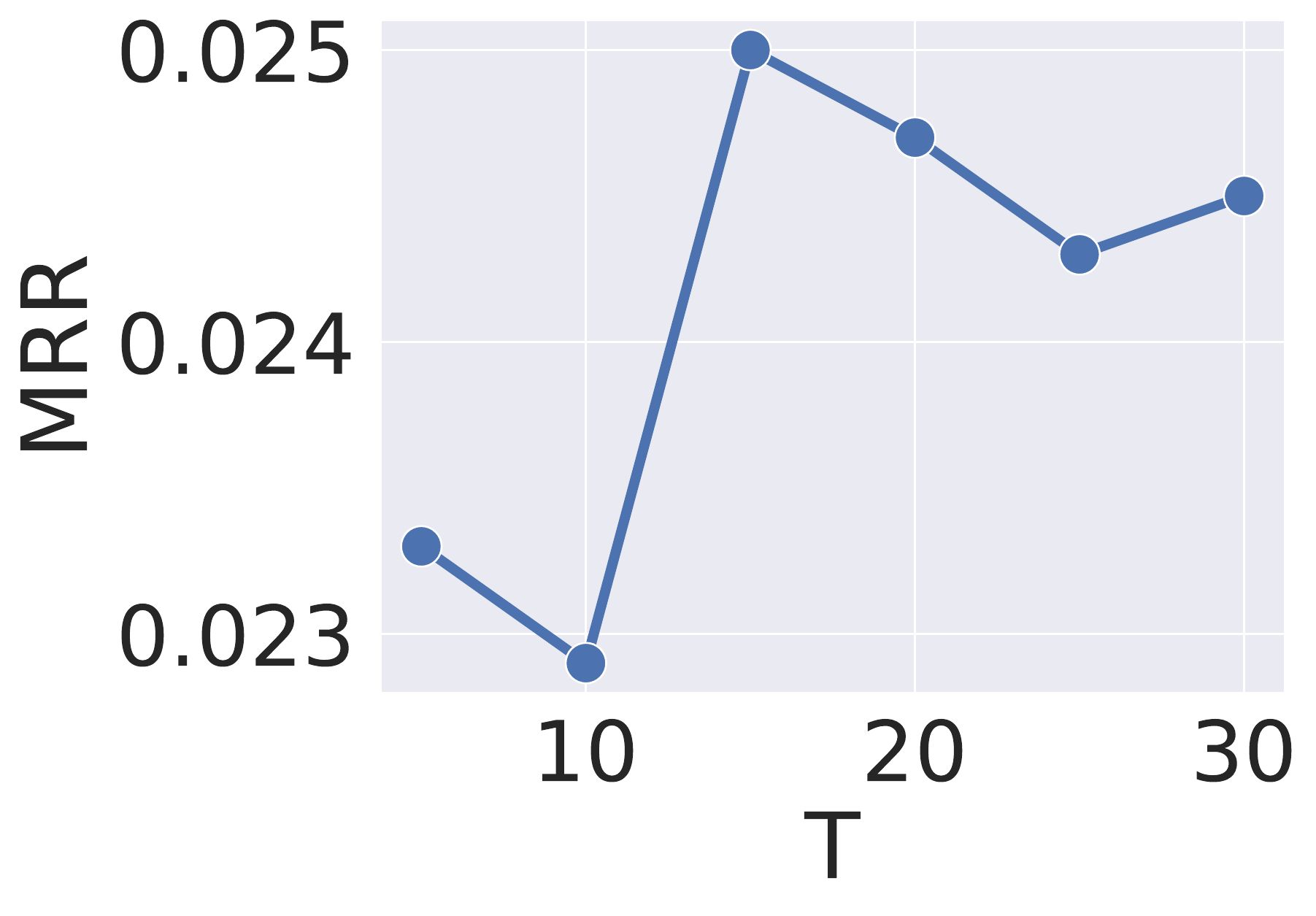}
        \caption{Tools}
        \label{fig:subfigure-c}
    \end{subfigure}
    \hfill
    \begin{subfigure}{0.24\textwidth}
        \centering
        \includegraphics[width=\linewidth]{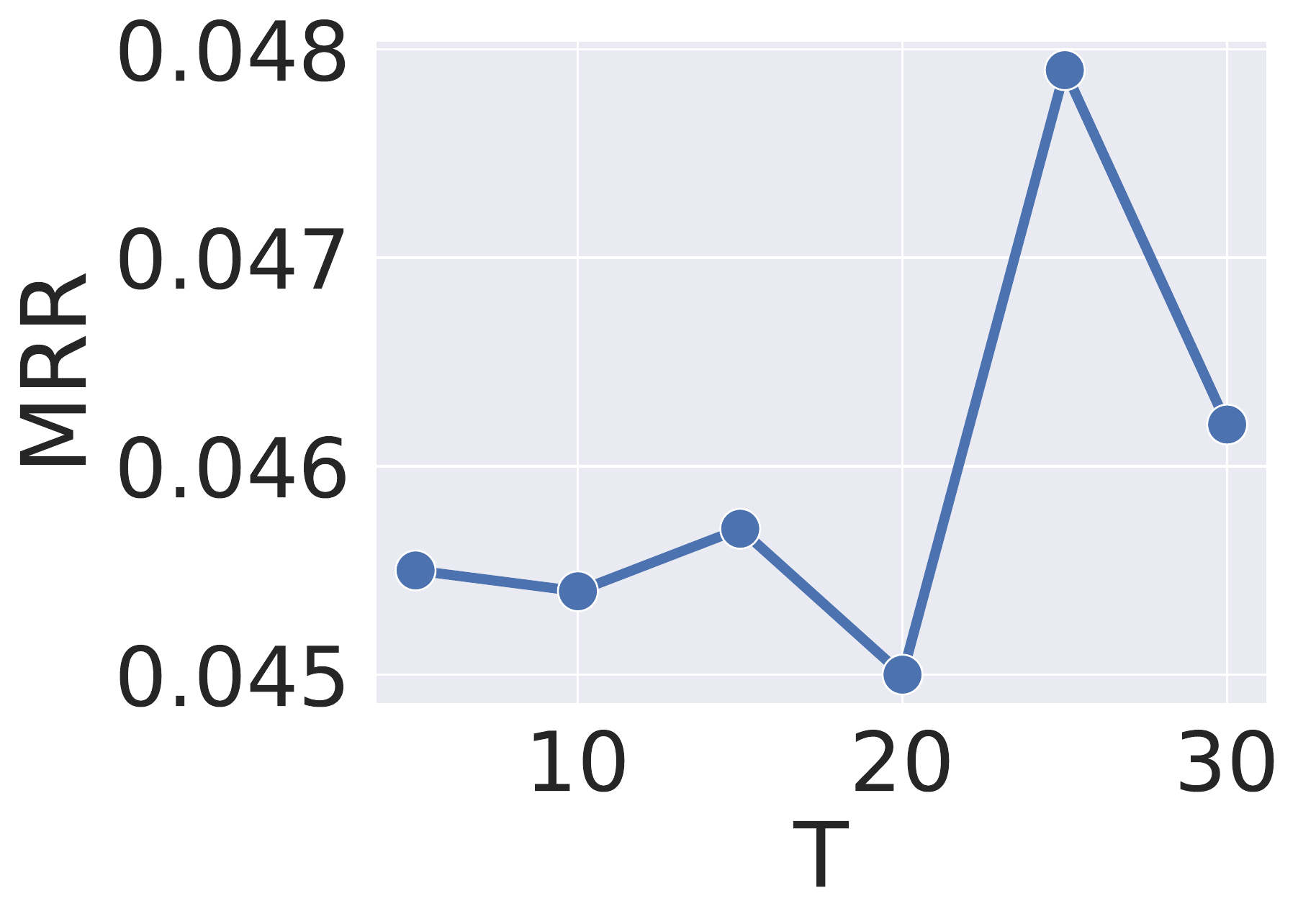}
        \caption{Toys}
        \label{fig:subfigure-d}
    \end{subfigure}
    \vspace{-2mm}
    \caption{The evaluation of \modelname on MRR through four datasets with different maximum diffusion step $T$.}
    \label{fig:T}
\end{figure*}

\begin{figure*}[h]
    \centering
    \begin{subfigure}{0.24\textwidth}
        \centering
        \includegraphics[width=\linewidth]{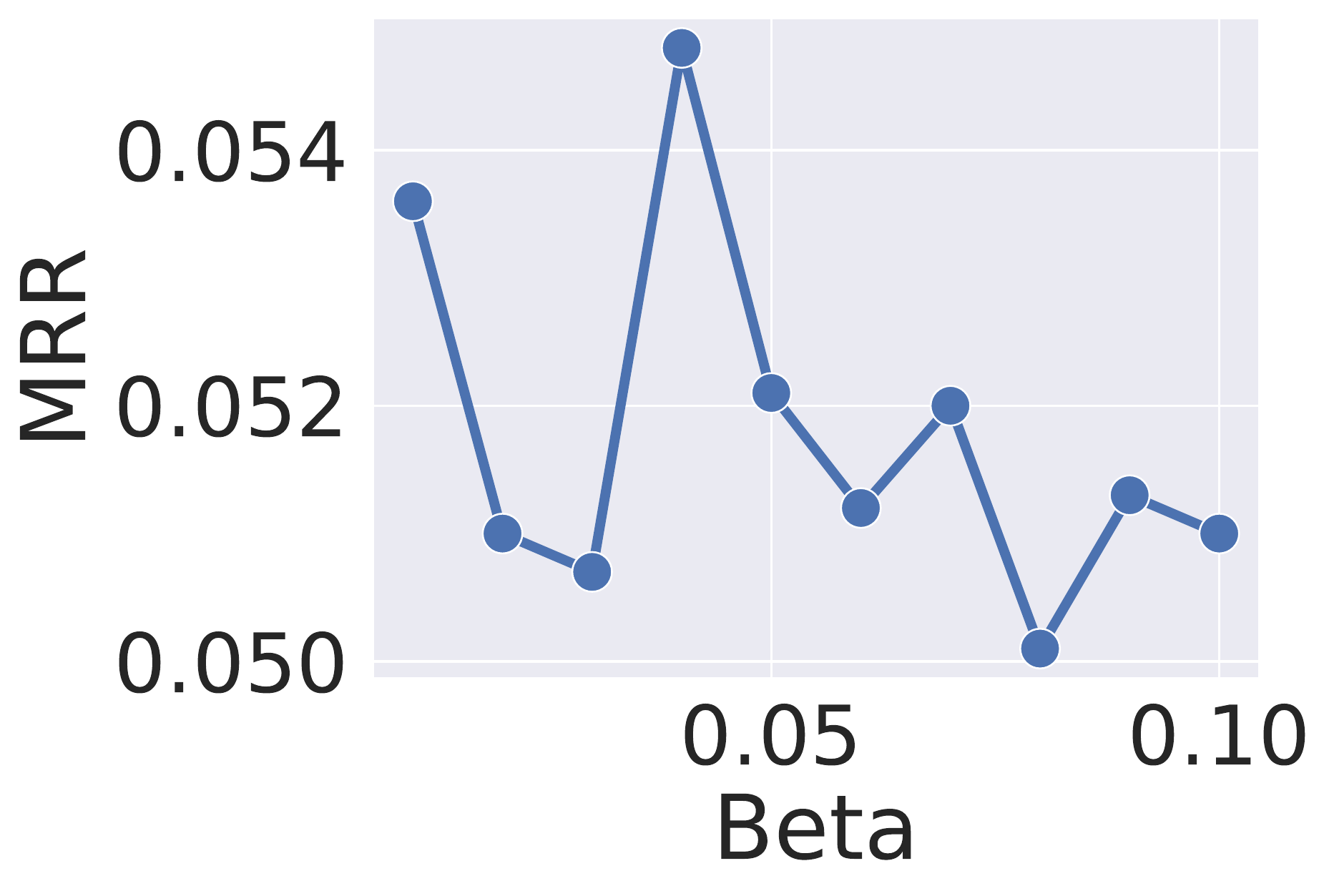}
        \caption{Office}
        \label{fig:subfigure-a}
    \end{subfigure}
    \hfill
    \begin{subfigure}{0.24\textwidth}
        \centering
        \includegraphics[width=\linewidth]{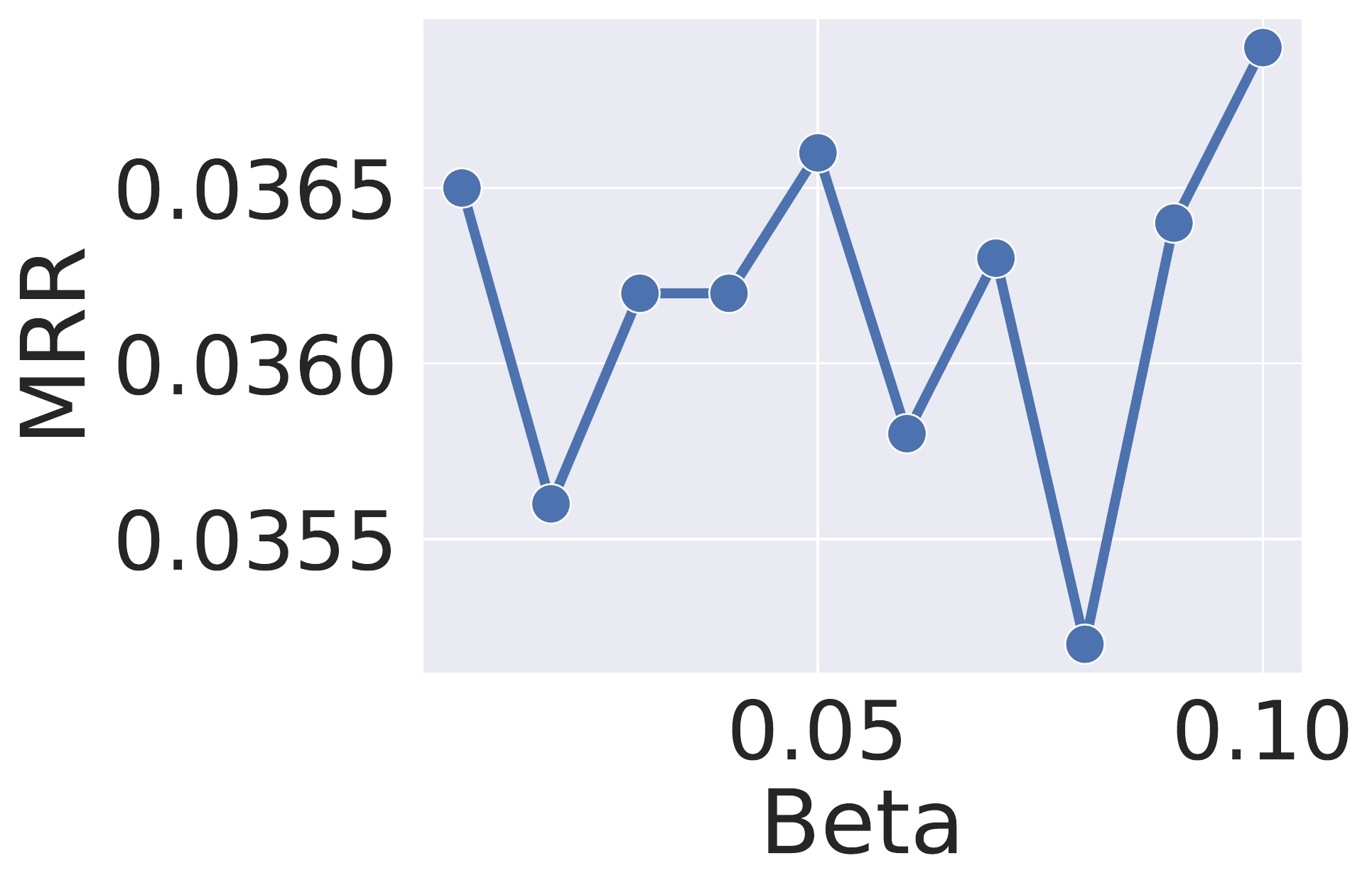}
        \caption{Beauty}
        \label{fig:subfigure-b}
    \end{subfigure}
    \hfill
    \begin{subfigure}{0.24\textwidth}
        \centering
        \includegraphics[width=\linewidth]{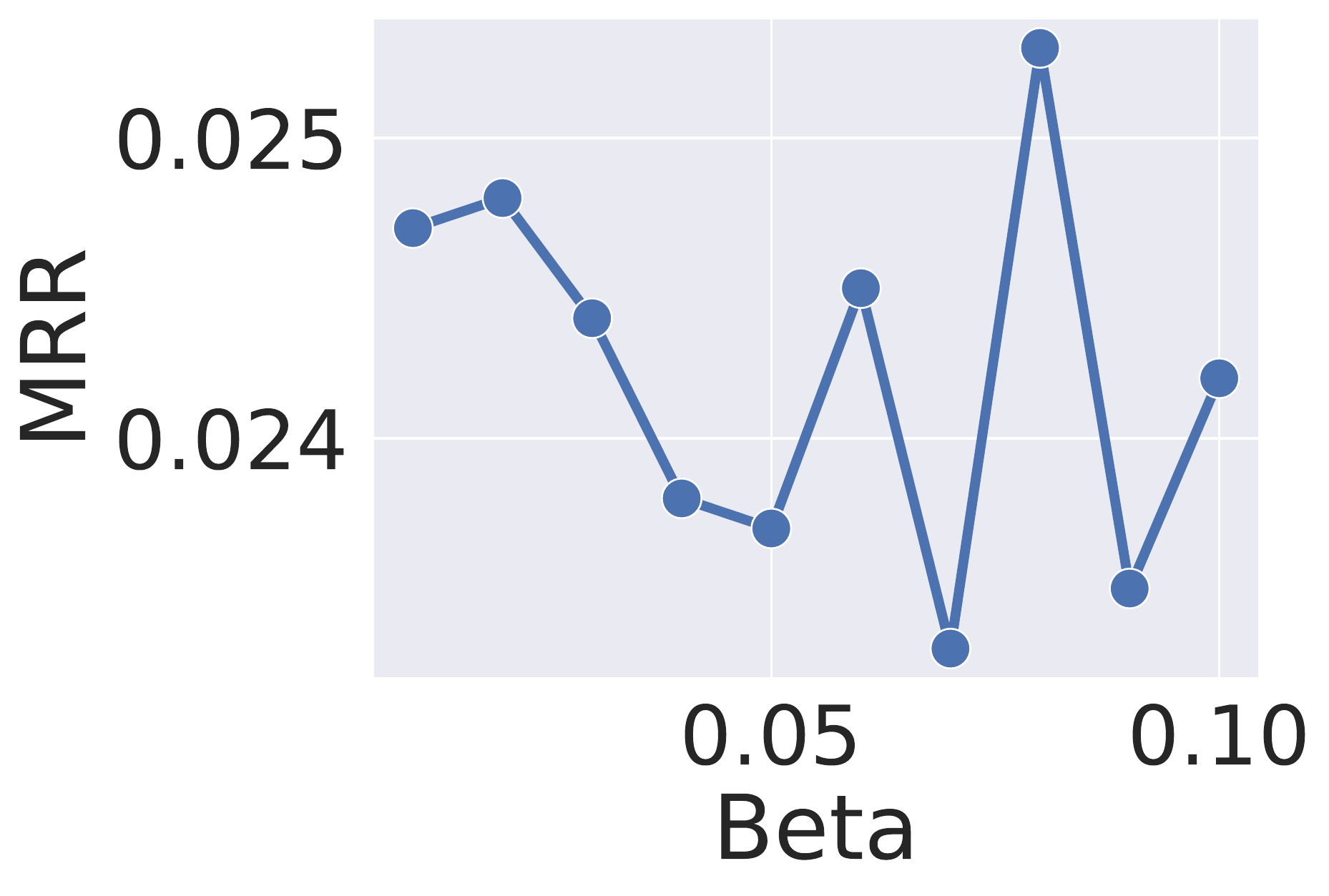}
        \caption{Tools}
        \label{fig:subfigure-c}
    \end{subfigure}
    \hfill
    \begin{subfigure}{0.24\textwidth}
        \centering
        \includegraphics[width=\linewidth]{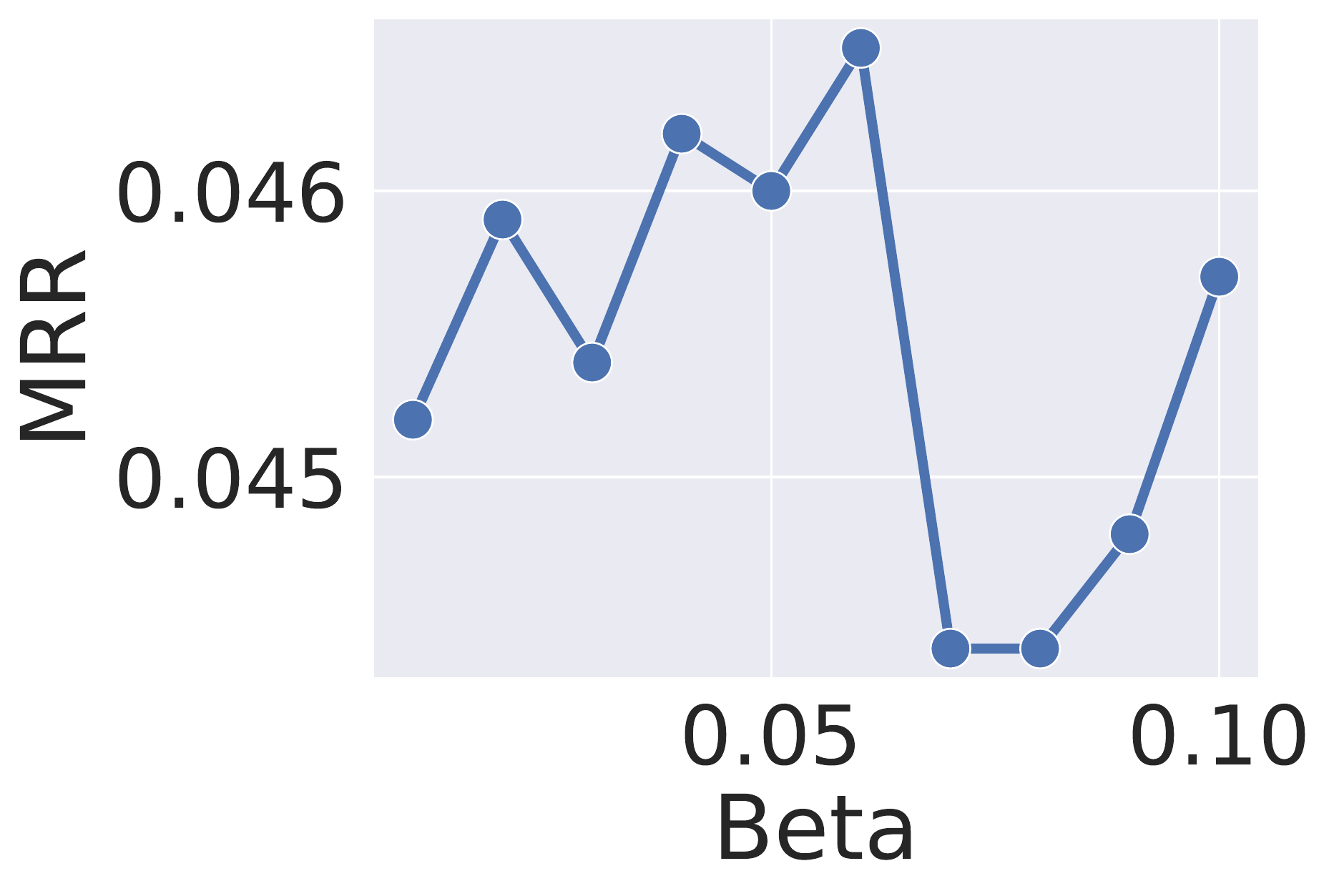}
        \caption{Toys}
        \label{fig:subfigure-d}
    \end{subfigure}
    \vspace{-2mm}
    \caption{The evaluation of \modelname on MRR through four datasets with different maximum noise schedules $\beta$.}
    \label{fig:beta}
\end{figure*}

In this section, we investigate the sensitivities of \modelname's hyperparameters. Due to space constraints, we focus on reporting the experimental results for key hyperparameters, including maximum diffusion step and maximum noise schedule. As depicted in Figure~\ref{fig:T}, the optimal maximum diffusion steps are $10$, $15$, $25$, and $30$ for Office, Tools, Toys, and Beauty datasets, respectively. This implies that the optimal diffusion step for each dataset is related to their item count. Specifically, the Beauty dataset has the highest number of items, making it more challenging for \modelname to learn meaningful item embeddings. Consequently, a greater number of diffusion steps are required to obtain more refined intermediate states of item representations. Additionally, we present the experimental results for \modelname with varying noise schedules in Figure~\ref{fig:beta}. We observe that the optimal maximum noise levels for Office, Toys, Tools, and Beauty are $0.04$, $0.06$, $0.08$, and $0.1$, respectively. A possible explanation for this pattern could be the combined influence of sequence length and item count. Longer sequences may inherently contain noisy interactions, thus necessitating less added noise. The maximum noise schedule for Office, Toys, and Tools aligns with this observation. However, the Beauty dataset, with the largest number of items, may demand more granular learning, thereby requiring higher levels of added noise to the sequences.
\vspace{-2mm}

\vspace{-2mm}

\subsection{Study the effect on Length and Frequency}

\begin{figure*}[h]
    \centering
    \begin{subfigure}{0.24\textwidth}
        \centering
        \includegraphics[width=\linewidth]{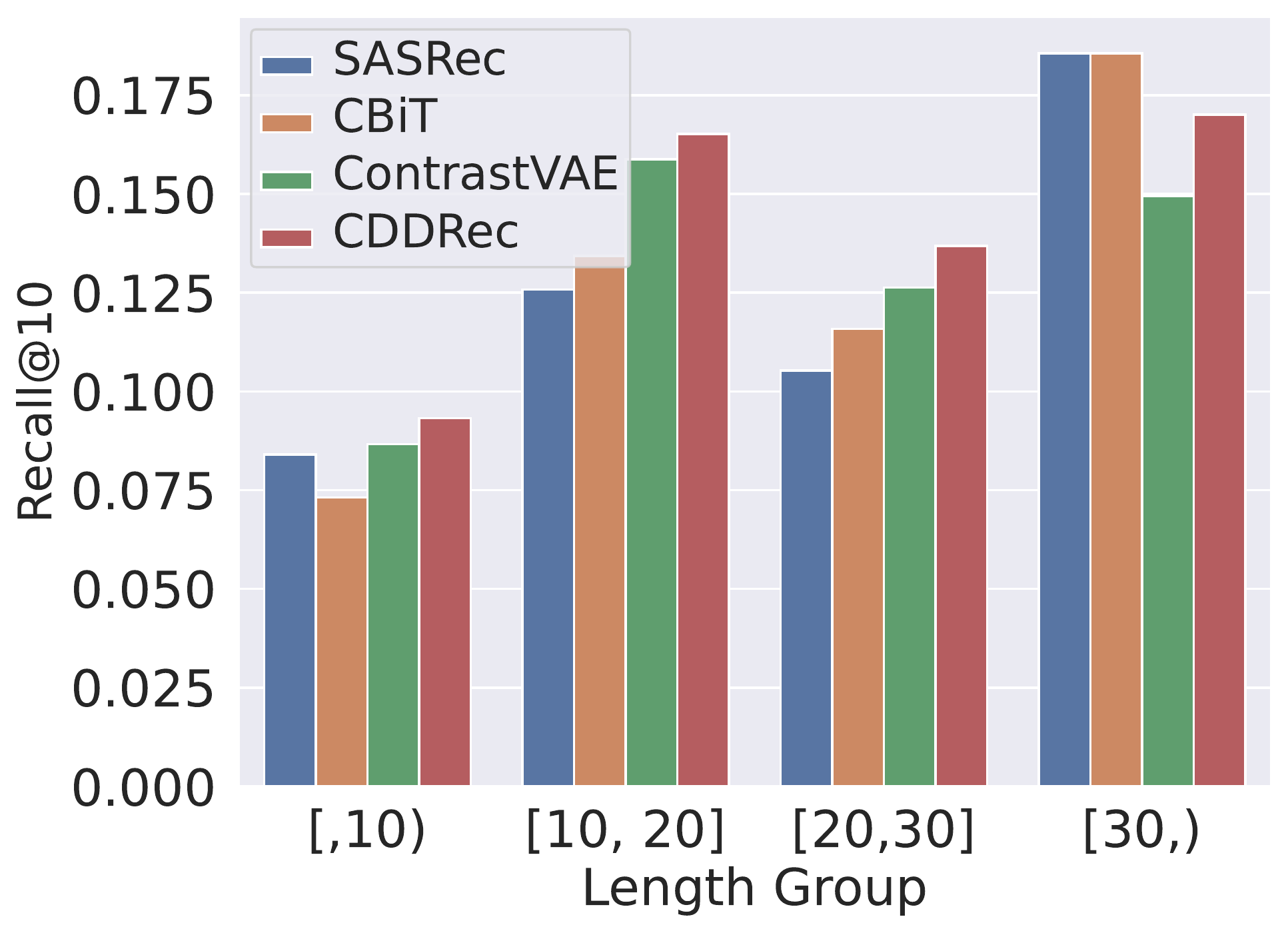}
        \caption{Length}
        \label{fig:subfigure-a}
    \end{subfigure}
    \hfill
    \begin{subfigure}{0.24\textwidth}
        \centering
        \includegraphics[width=\linewidth]{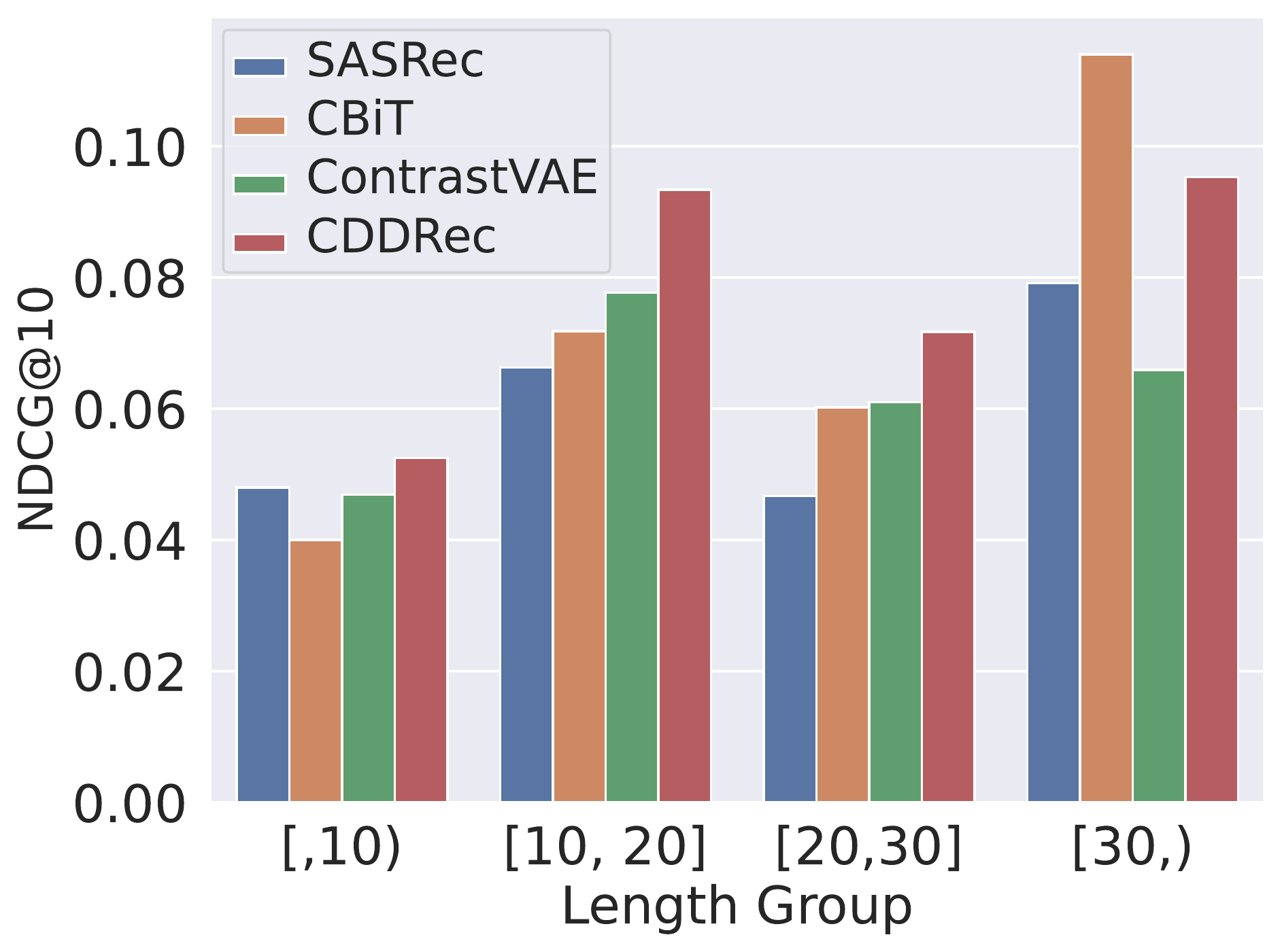}
        \caption{Length}
        \label{fig:subfigure-b}
    \end{subfigure}
    \hfill
    \begin{subfigure}{0.24\textwidth}
        \centering
        \includegraphics[width=\linewidth]{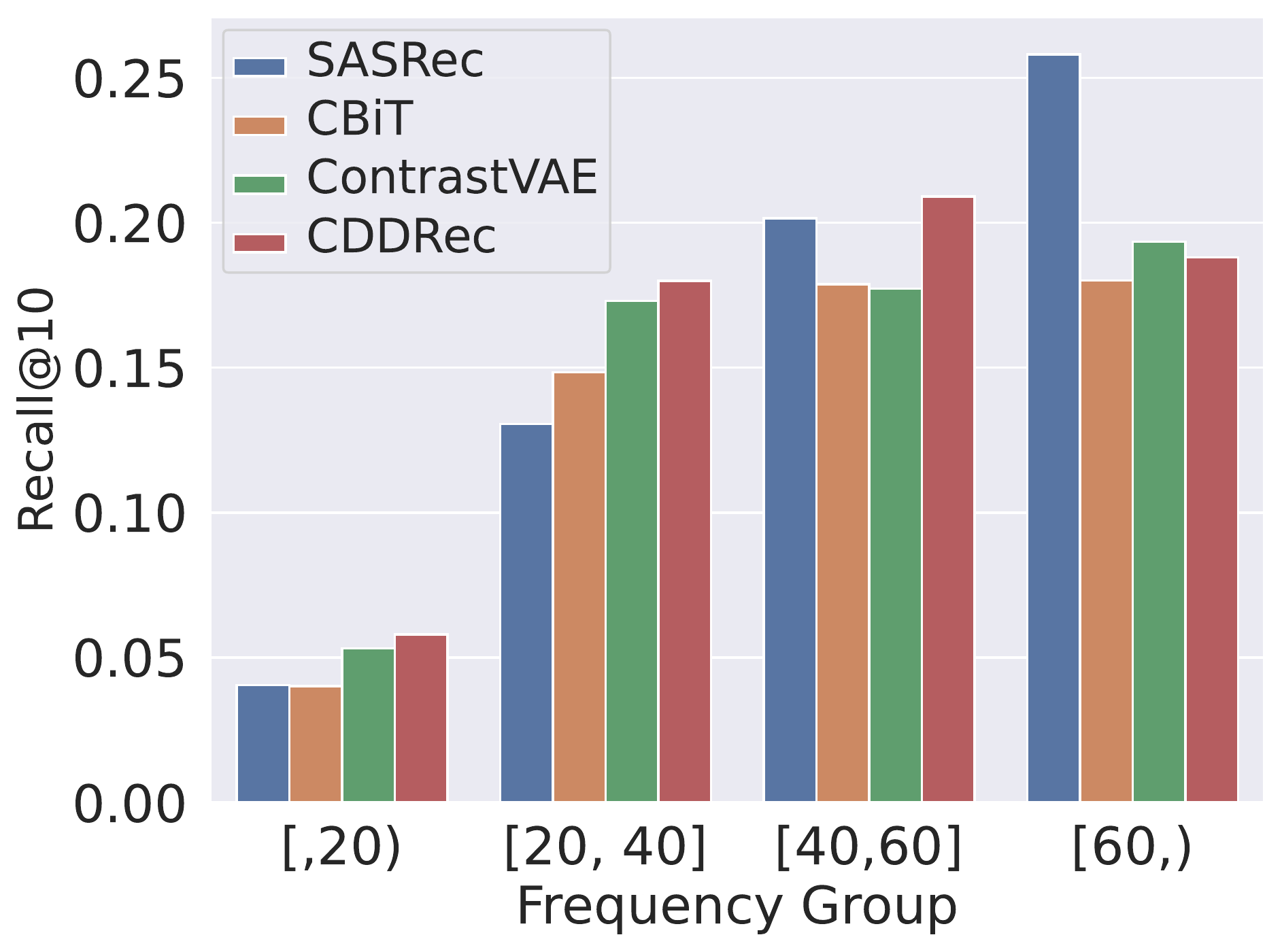}
        \caption{Frequency}
        \label{fig:subfigure-c}
    \end{subfigure}
    \hfill
    \begin{subfigure}{0.24\textwidth}
        \centering
        \includegraphics[width=\linewidth]{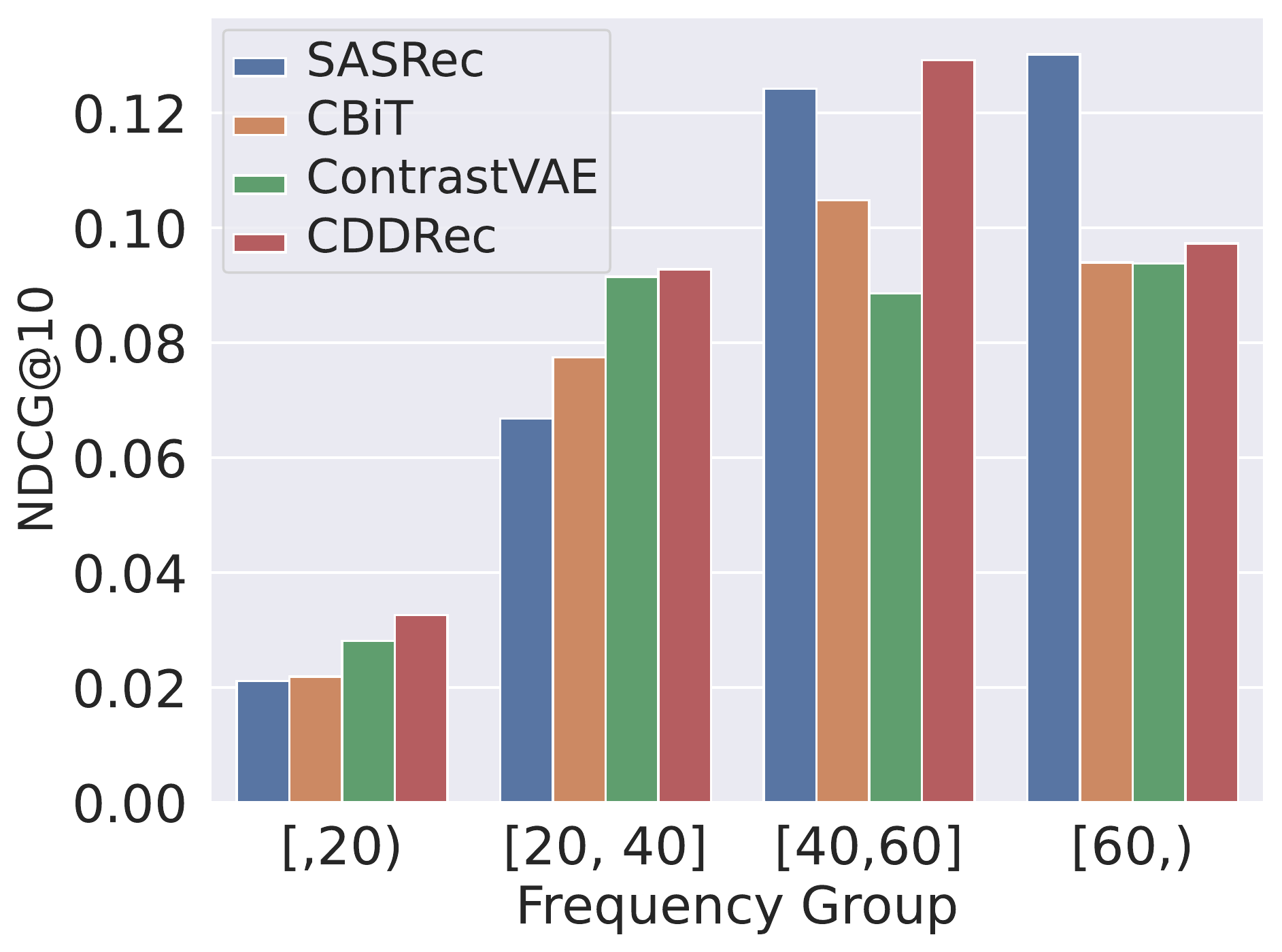}
        \caption{Frequency}
        \label{fig:subfigure-d}
    \end{subfigure}
    \vspace{-2mm}
    \caption{The comparison between \modelname and baseline models on subgroups of Office dataset.}
    \label{fig:len_freq}
\end{figure*}

In this section, we present a comparative analysis of the performance of \modelname and baseline methods CBiT, ContrastVAE, and SASRec on the Office dataset. We first compute the training sequence length and target item frequency based on their occurrence in the training dataset. Subsequently, we assess the Recall@10 and NDCG@10 metrics. We report the dataset statistics concerning sequence length and item frequency in Table~\ref{tab:len_freq_ana}, and present the performance comparison in Fig.~\ref{fig:len_freq}.

\paragraph{Discussion on Sequence Length.} Our analysis yields the following observations: Firstly, \modelname consistently surpasses the baseline models for shorter sequences across all metrics, particularly for sequences with a length of fewer than 30 items. This can be attributed to \modelname's diffusion and denoising mechanism, which introduces Gaussian Noise to the predicted sequence representation and target sequence embeddings. Shorter sequences are more susceptible to the impact of noisy interactions. By explicitly adding noise to sequences, the model exhibits increased robustness to noisy interactions in the dataset. As 95\% of the Office dataset consists of short sequences with fewer than 30 interactions, the improvement in performance for such sequences significantly contributes to our model's overall superiority.

\paragraph{Discussion on Item Frequency.} With respect to sequence subsets featuring different target item frequencies, \modelname outperforms the baseline models for sequences wherein target items exhibit lower frequencies in the training dataset. A possible explanation for this phenomenon is that \modelname learns multi-step intermediate states of transitions from sequence embeddings to target item embeddings. Low-frequency items may serve as intermediate transitions, enabling the model to recognize latent transitions from frequent to infrequent items.
\vspace{-2mm}

\vspace{-2mm}
\begin{table*}[h]
    \centering
    \caption{Number of sequences end at items of different frequency groups and with different lengths.}
    \vspace{-2mm}
    \setlength{\tabcolsep}{2.6mm}{
    \begin{tabular}{c|cccc|cccc}
         \hline
         &\multicolumn{4}{c|}{Frequency} & \multicolumn{4}{c}{Length}\\
         Dataset & [$\leq$20] & [20, 40] & [40, 60] & [$\geq$60] & [$\leq$10] & [10, 20] & [20, 30] & [$\geq$30]  \\
         \hline
         Office & 2862 & 1011 & 660 & 372 & 3488 & 938 & 285 & 194 \\
         Beauty & 14271 & 3475 & 1878 & 2739 & 18816 & 2528 & 568 & 451 \\
         Tools & 12049 & 2487 & 921 & 1181 & 14516 & 1659 & 283 & 180 \\
         Toys & 13711 &  3379 & 1109 & 1213 & 16570&  2034 & 446 & 362 \\
         \hline
    \end{tabular}
    }
    \label{tab:len_freq_ana}
\end{table*}

\subsection{Study of Smoothness of Ranking Prediction}

As previously mentioned, traditional generative models often produce over-smoothed outputs. Intuitively, in the context of sequential recommendation, generative models may predict over-smoothed ranking scores across candidate items. To support this intuition, we carry out experiments comparing the average absolute percentage change (Avg.Change) of the top 40 ranking scores between ContrastVAE and \modelname. The metric is defined as follows:
\begin{equation}
    \text{Avg.Change} = \sum_{i=1}^N \frac{1}{N-1} \frac{|rank_{i+1}-rank_{i}|}{rank_i} \times 100
\end{equation}, where $rank$ is the ranking score vector calculated with the dot-product between predicted item embeddings and the candidate item embeddings. 

\begin{figure*}[h]
    \centering

    \hfill
    \begin{subfigure}{0.3\textwidth}
        \centering
        \includegraphics[width=\linewidth]{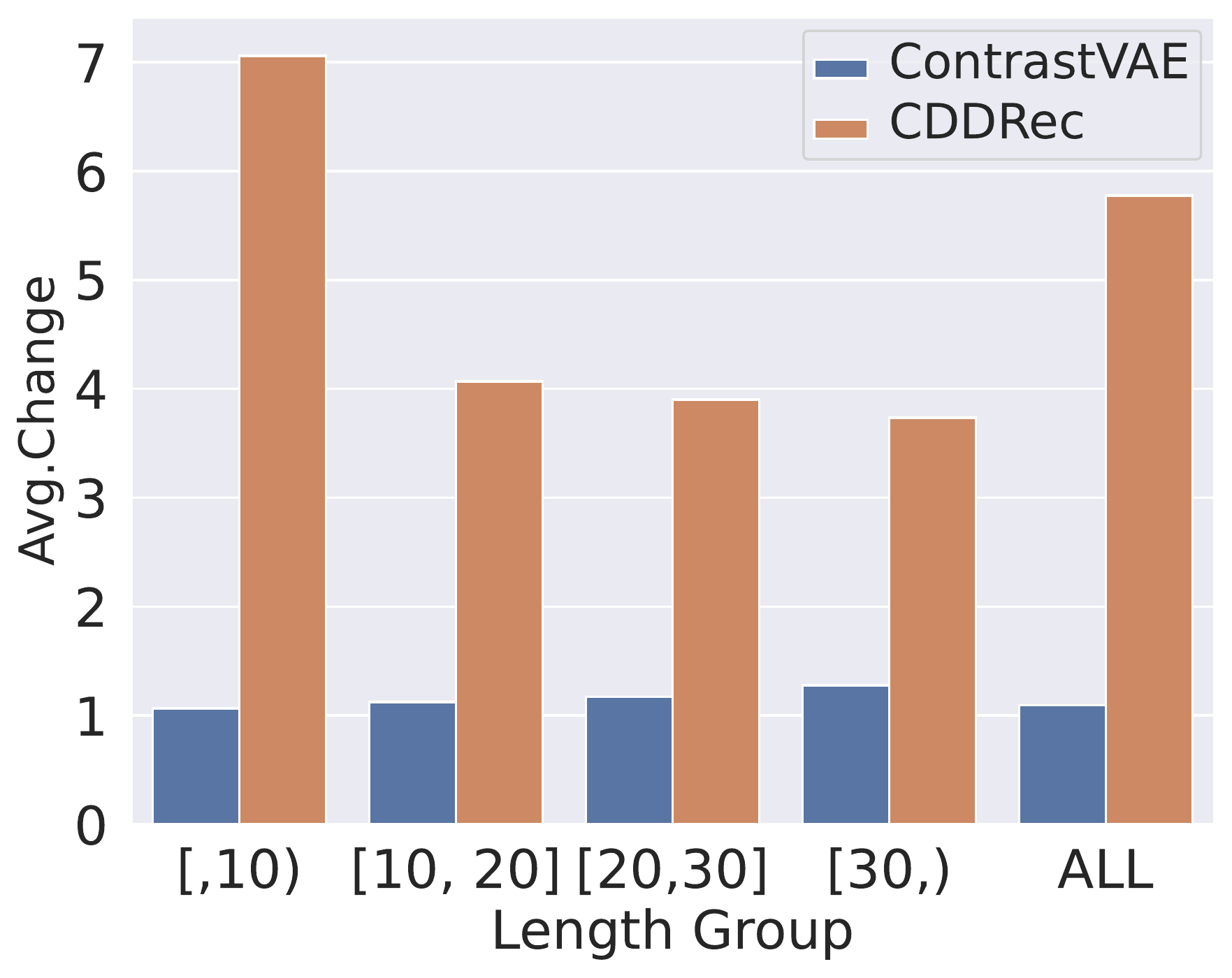}
        \caption{Length}
        \label{fig:subfigure-b}
    \end{subfigure}
    \hfill
    \begin{subfigure}{0.3\textwidth}
        \centering
        \includegraphics[width=\linewidth]{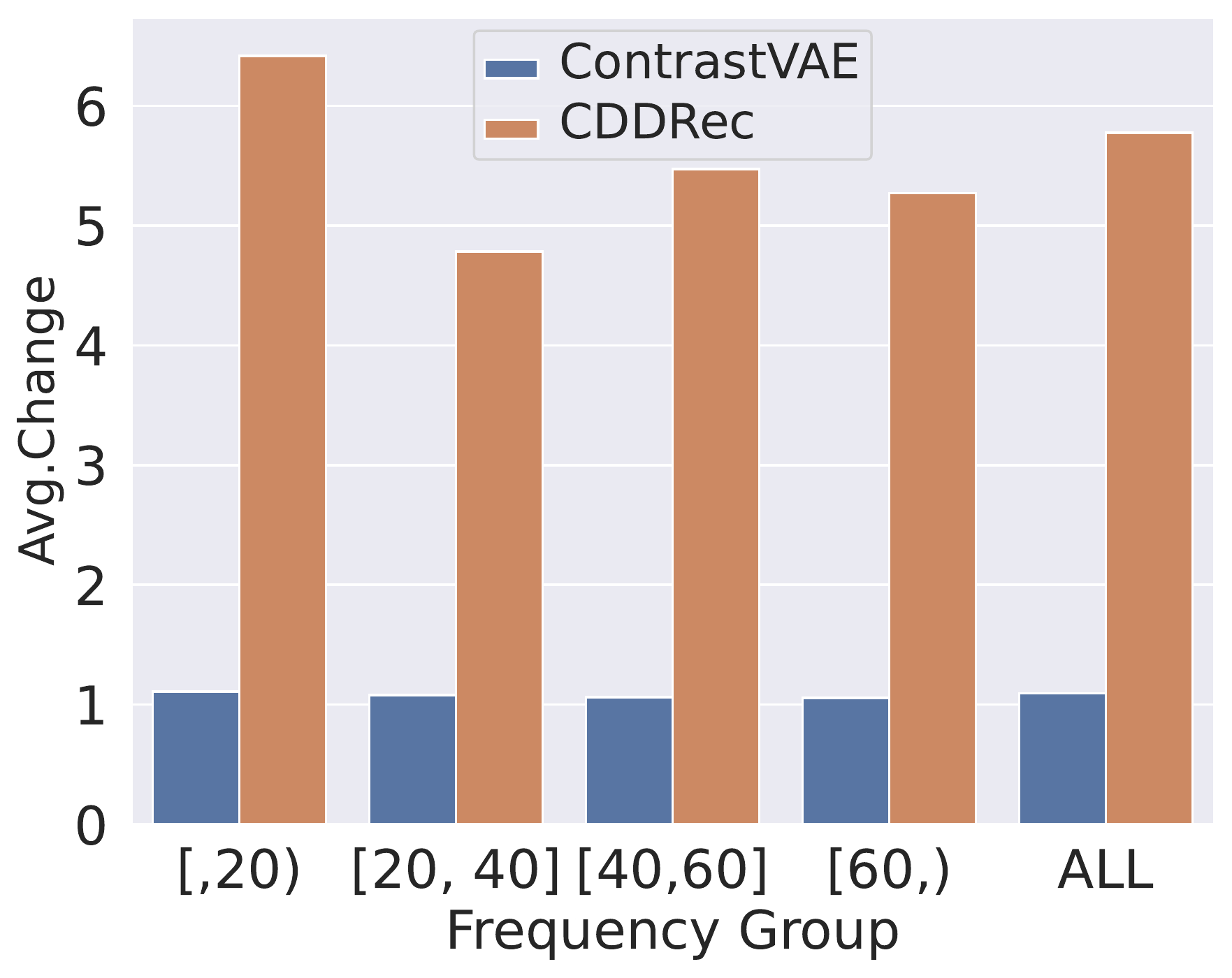}
        \caption{Frequency}
        \label{fig:subfigure-c}
    \end{subfigure}
    \hfill
        \begin{subfigure}{0.3\textwidth}
        \centering
        \includegraphics[width=\linewidth]{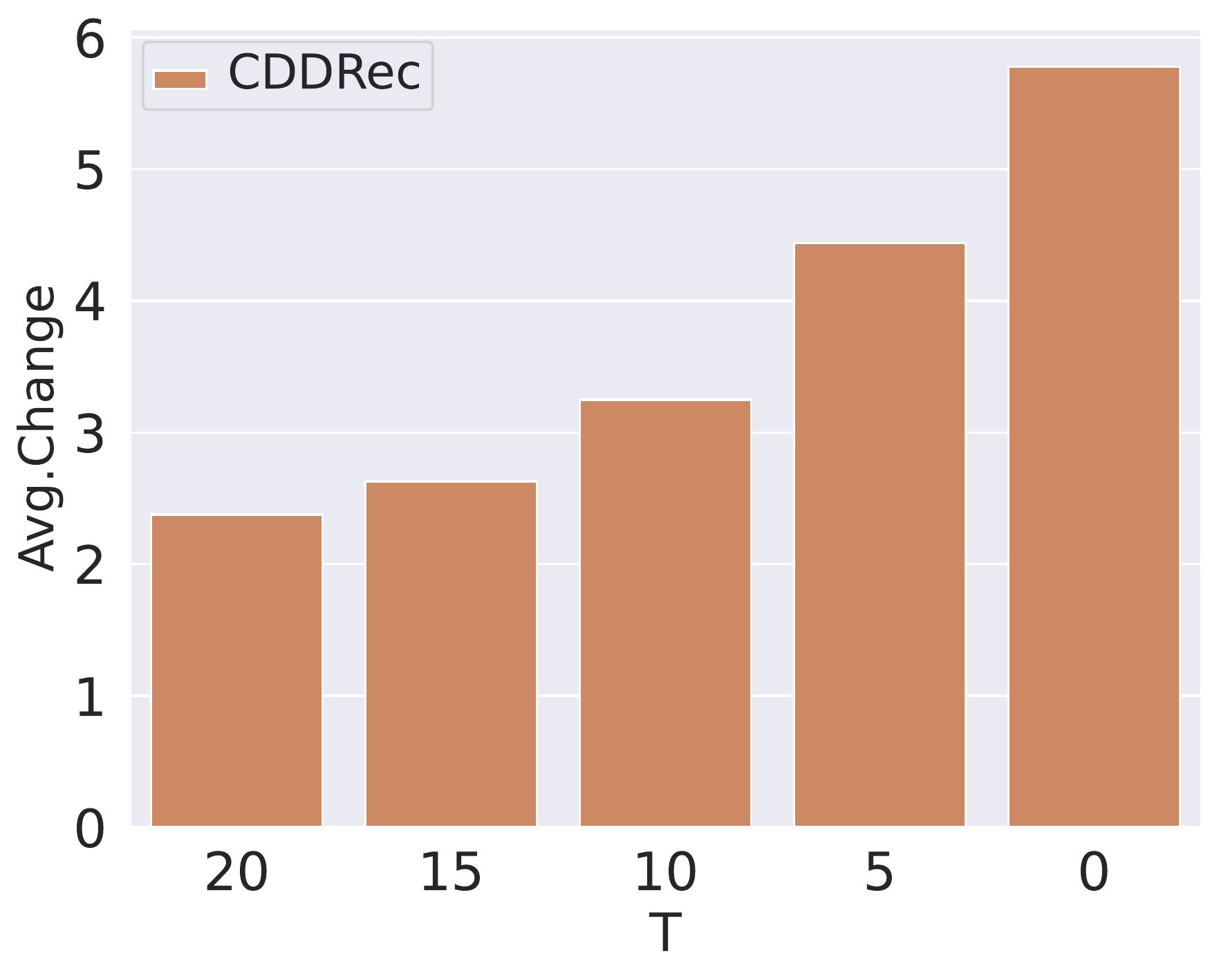}
        \caption{Denoising Step T}
        \label{fig:subfigure-c}
    \end{subfigure}
    \hfill
    \caption{The Avg.Change for \modelname and ContrastVAE across various subset sequences and denoising step T on Office dataset.}
    \label{fig:change}
\end{figure*}

We utilize this metric to evaluate the descending speed of the model's ranking prediction, which can reflect the smoothness of the ranking prediction. The results are presented in Fig.~\ref{fig:change}. In general, \modelname exhibits greater confidence in its ranking predictions, as evidenced by the larger average percentage change compared to ContrastVAE for both overall evaluation and sequence subsets. ContrastVAE is inclined to provide more similar ranking predictions among the top 40 candidates. Interestingly, the Avg.Change of \modelname decreases with increasing sequence length, indicating that the model is more uncertain about recommendations for longer sequences. Concerning item frequency, \modelname displays a larger average percentage change for sequences featuring infrequent target items.
We also evaluate the Avg.Change with respect to the denoising step. We calculate the ranking score vectors on candidate items using intermediate item embedding predictions from different denoising steps and report their Avg.Change in Fig.~\ref{fig:change}(c). One interesting observation is that the Avg.Change increases with the denoising step. At the beginning of the denoising phase (T=20), the model is more uncertain about ranking predictions, gradually gaining clarity (T=0) as the denoising phase progressively removes noise from the sequence predictions.

\section{Conclusion}
In conclusion, we introduce \modelname, featuring a cross-attentive conditional denoising decoder that utilizes the denoising step indicator as the query information and the sequence embedding as the key and value information to endow the model with conditional autoregressive generation capabilities. Additionally, we propose the recommendation optimization paradigm for \modelname, enabling the model to generate high-fidelity sequence/item representations and provide high-quality ranking predictions. We conduct comprehensive experiments that indicate that \modelname outperforms state-of-the-art methods.

\bibliographystyle{ACM-Reference-Format}
\bibliography{references}
\end{document}